\documentclass[preprint,12pt]{elsarticle}
\usepackage[english]{babel}
\usepackage{xcolor}
\usepackage{float}
\usepackage{pifont}
\usepackage[caption=false]{subfig}%
\usepackage{booktabs} 
\usepackage{multirow}
\usepackage{xspace}
\usepackage{enumitem}
\usepackage{balance}
\usepackage[multiple]{footmisc}
\usepackage{hyperref}
\usepackage{booktabs}
\usepackage{caption}
\usepackage[autostyle]{csquotes}
\usepackage{listings}
\usepackage{tikz}
\usepackage{amsmath,amssymb,amsfonts}
\usepackage{algorithmic}

\usepackage{pifont}
\usepackage{graphicx}
\usepackage{textcomp}
\usepackage{adjustbox}
\usepackage{lscape}
\usepackage{tabularx}
\usepackage{color}
\usepackage[newcommands]{ragged2e}
\usepackage{colortbl}
\usepackage{array}

\newcolumntype{C}[1]{>{\centering\arraybackslash\hspace{0pt}}p{#1}}
\def\BibTeX{{\rm B\kern-.05em{\sc i\kern-.025em b}\kern-.08em
   T\kern-.1667em\lower.7ex\hbox{E}\kern-.125emX}}
  
\journal{Computer Networks}  
\begin{document}
\begin{frontmatter}

\title{Evaluating Federated Learning for Intrusion Detection in Internet of Things: Review and Challenges}

\author[inst1]{Enrique Mármol Campos}
\author[inst1]{Pablo Fernández Saura}
\author[inst1]{Aurora González-Vidal}
\author[inst2]{José L. Hernández-Ramos}
\author[inst1]{Jorge Bernal Bernabe}
\author[inst2]{Gianmarco Baldini}
\author[inst1]{Antonio Skarmeta}

\affiliation[inst1]{organization={University of Murcia, Department of Information and Communication Engineering, Spain (e-mail: \{enrique.marmol, pablo.fernandezs2, aurora.gonzalez2, jorgebernal, skarmeta\}@um.es)}}

\affiliation[inst2]{organization={European Commission, Joint Research Centre, Ispra 21027, Italy (e-mail: \{jose-luis.hernandez-ramos, gianmarco.baldini\}@ec.europa.eu)}}


\begin{abstract}
The application of Machine Learning (ML) techniques to the well-known intrusion detection systems (IDS) is key to cope with increasingly sophisticated cybersecurity attacks through an effective and efficient detection process. In the context of the Internet of Things (IoT), most ML-enabled IDS approaches use centralized approaches where IoT devices share their data with data centers for further analysis. To mitigate privacy concerns associated with centralized approaches, in recent years the use of Federated Learning (FL) has attracted a significant interest in different sectors, including healthcare and transport systems. However, the development of FL-enabled IDS for IoT is in its infancy, and still requires research efforts from various areas, in order to identify the main challenges for the deployment in real-world scenarios. In this direction, our work evaluates a FL-enabled IDS approach based on a multiclass classifier considering different data distributions for the detection of different attacks in an IoT scenario. In particular, we use three different settings that are obtained by partitioning the recent ToN\_IoT dataset according to IoT devices' IP address and types of attack. Furthermore, we evaluate the impact of different aggregation functions according to such setting by using the recent IBMFL framework as FL implementation. Additionally, we identify a set of challenges and future directions based on the existing literature and the analysis of our evaluation results.
\end{abstract}

\begin{highlights}
    \item Analysis of existing of FL-enabled IDS approaches for IoT based on a set of identified criteria.
    \item Partitioning of the recent ToN\_IoT dataset to evaluate the impact of data distribution in a multi-class classifier for detecting specific types of attacks.
    \item Quantitative analysis of the impact of non-iid data considering different aggregation methods and training rounds by using the recent IBMFL implementation.
    \item Definition of the main challenges and future trends to be considered in the coming future for the development of FL-enabled IDS for IoT scenarios.
\end{highlights}

\end{frontmatter}

\section{Introduction}\label{sec:intro}
Nowadays, the constant development and deployment of Internet of Things (IoT) technologies is increasing the attack surface of physical devices that could be potentially exploited by malicious entities \cite{neshenko2019demystifying}. Well-known attacks, such as the Mirai botnet and recent variants \cite{pour2020data}, demonstrate the need to strengthen IoT devices’ security in order to protect large-scale IoT-enabled systems. Due to the development of such increasingly sophisticated attacks, in recent years the use of machine learning (ML) techniques has been widely considered for the detection and mitigation of these attacks in IoT scenarios. Indeed, the application of ML techniques has been proposed in recent works to improve the detection capabilities of the well-known intrusion detection systems (IDS) through the application of diverse techniques (e.g., neural networks) to infer potential attacks based on the analysis of network traffic \cite{da2019internet}.
Despite the advantages provided by the application of ML techniques to enhance IDS approaches (e.g., in terms of attack detection accuracy), a main limitation of current approaches is that they are based on a centralized training process in which a single entity receives the network traffic data from different devices to train a certain ML model. Therefore, such component has access to the whole network traffic derived from the communication of the different devices participating in the training process and also devices' local data. This aspect could lead to privacy issues, which could be exacerbated in IoT scenarios due to the amount and sensitivity of the information exchanged through certain devices, such as wearable or eHealth systems \cite{ding2019survey}; therefore, decentralised solutions to manage data are of great importance \cite{iggena2021iotcrawler}.

To address the limitations of traditional centralized ML approaches, Federated Learning (FL) was proposed in 2016 \cite{mcmahan2017communication} as a collaborative learning approach in which end devices (a.k.a clients or parties) do not share their data, but only partial updates of a global model that are aggregated by a central entity (a.k.a aggregator or coordinator). Therefore, the use of FL is intended to improve users’ privacy, since the data of their devices is never shared with other entities. In general, an FL scenario is characterized by a large number of client devices with a variable amount and distribution of data. Indeed, real-life scenarios are usually based on non-independent and identically distributed (non-iid) data \cite{zhao2018federated}. For example, in the case of an IDS deployed on a certain network, some target devices could have traffic associated with several kinds of attacks (e.g., DoS or port scanning), while other devices could only have traffic related to their intended operation. 
The development of FL-enabled IDS approaches in the context of IoT scenarios has attracted an increasing interest in recent years \cite{DIOT,FLMimicIDS, LocKedge}. However, most of the proposed approaches are based on unrealistic data distributions among the parties, inappropriate datasets and settings (e.g., \cite{BFL-CIDS}), or they use binary classification approaches, in which traffic data is only classified as attack or benign \cite{rey2021federated}. Consequently, it is hard for cybersecurity practitioners to come up with the most challenging aspects derived from the application of FL to enhance IDS approaches in IoT.

To fill this gap, our work provides a comprehensive evaluation on the use of FL for IDS in IoT by considering the impact of non-iid data. In particular, we evaluate the behavior of FL by considering different data distributions, training rounds and aggregation methods. For this purpose, we use the ToN\_IoT dataset \cite{booij2021ton_iot, alsaedi2020ton_iot}, which has recently been proposed for IoT and Industrial IoT scenarios considering sensor data manipulation attacks, in addition to several network attacks. We propose three scenarios based on different partitions and processing of the ToN\_IoT dataset: in the first setting, network flows are split according to their destination IP address; the second scenario is balanced according to the types of attacks among the clients; then, a hybrid approach is considered as third setting, in which we find a compromise between the balance of attack types and the destination IP address by means of the Shannon entropy \cite{bonachela2008entropy}. These three configurations are publicly available at \cite{datasets}. Then, we evaluate such scenarios by using FedAvg \cite{li2019convergence} and Fed+ \cite{fed+paper} aggregation methods through the IBM framework for Federated Learning \textit{IBMFL} \cite{ludwig2020ibm}. Based on our evaluation results, and the analysis of the existing literature, we describe some of the main challenges for the development of FL-based IDS approaches to be deployed in IoT scenarios. Therefore, our work can be used as a reference for future research activities on the use of FL in this context. In summary, our contributions are:

\begin{itemize}
    \item Identification of the main aspects for the evaluation of FL-enabled IDS for IoT, and analysis of existing proposals according to such aspects.
    \item Partitioning of the recent ToN\_IoT dataset to create different data distributions among clients to evaluate its impact on the overall system accuracy.
    \item Quantitative analysis of the impact of non-iid data considering different aggregation methods and training rounds by using the recent IBMFL implementation.
    \item Usage of multi-class classification for differentiating specific types of attacks in the output.
    \item Definition of the main challenges and future trends to be considered in the future years for the development of FL-enabled IDS for IoT scenarios.
\end{itemize}

The structure of the paper is organized as follows. Section \ref{sec:background} provides an overview of FL and the main advantages derived from its application to IDS. In Section \ref{sec:related}, we describe and classify the existing research proposals on FL-enabled IDSs for IoT.  Section \ref{sec:methodology}  describes our methodology, including the aspects of the dataset partitioning, classification techniques and aggregation methods. Then, evaluation results are presented in Section \ref{sec:evaluation}. Based on such results and the analysis of existing literature, Section \ref{sec:challenges} highlights the main challenges for the development of FL-enabled IDS for IoT. Finally, Section \ref{sec:conclusions} concludes the paper with an outlook of potential future directions to be considered. 
\section{FL-enabled IDS for IoT scenarios}\label{sec:background}
Intrusion detection systems (IDS) have traditionally been considered as key components to protect ICT systems by identifying potential security attacks/threats derived from traffic monitoring and analysis. Although there are several classifications \cite{da2019internet} \cite{khraisat2019survey}, IDS approaches are usually categorized as signature and anomaly based systems. The former is based on pre-established network patterns and, consequently, it cannot be used to detect a new attack; the latter uses specific features of network traffic, so that a certain deviation on such network behavior can be considered as a potential attack. In recent years, the application of ML techniques to IDS has attracted a strong interest \cite{chapaneri2019comprehensive, liu2019machine} considering different approaches such as neural networks \cite{yin2017deep}, \cite{drewek2021survey} or clustering techniques \cite{liang2019industrial}. In the context of IoT, recent efforts have been proposed by considering specific IoT devices and technologies \cite{da2019internet}. Indeed, the use of Deep Learning (DL) techniques has been recently evaluated  through different types of neural networks for the detection of different attacks in such scenarios \cite{ferrag2021deep, ge2021towards, garcia2021distributed}.

\begin{figure*}[!htb]
    \centering
    \includegraphics[width=0.99\textwidth]{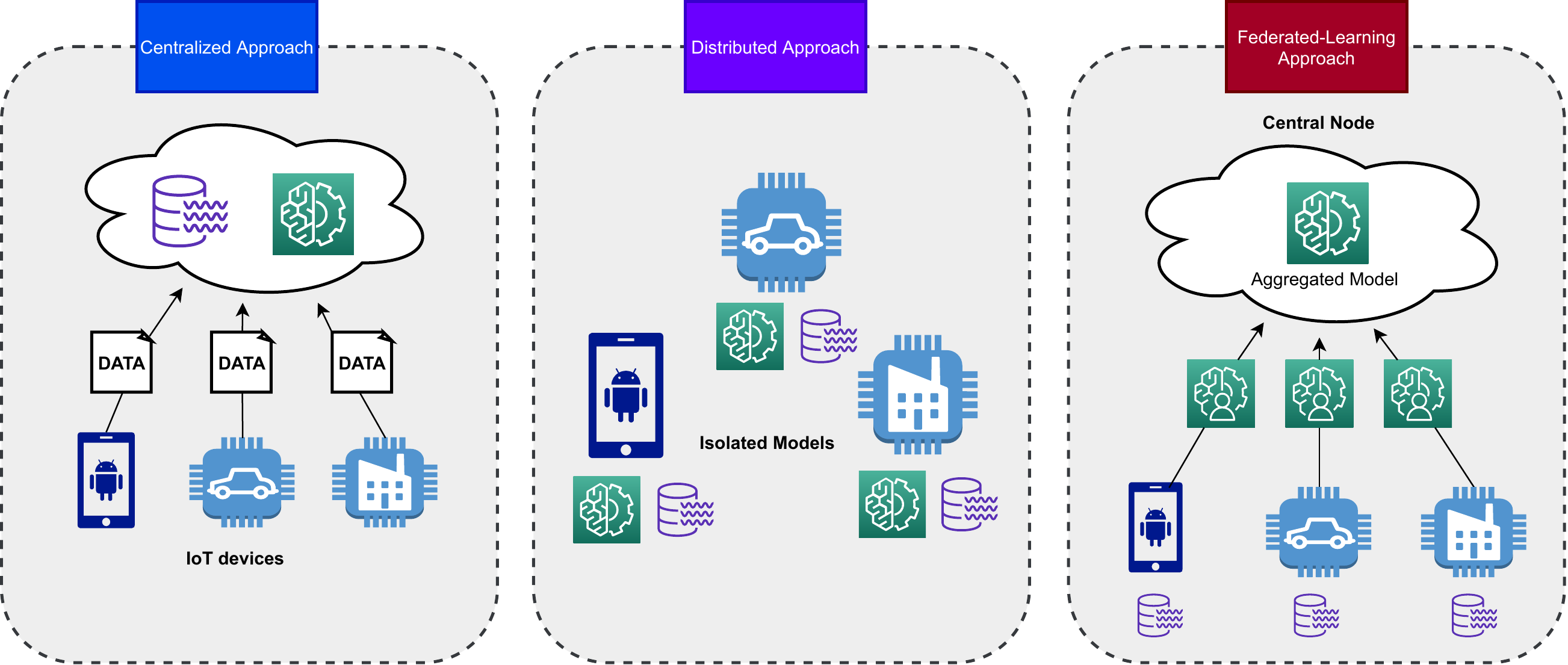}
    \caption{Comparison between centralized, distributed and federated learning approaches}
    \label{Fig:FLscheme}
\end{figure*}

Despite these efforts, most of the proposed IDS approaches for IoT are based on centralized approaches in which devices send their local data to data centers in the cloud or servers with considerable computing capabilities to be analyzed through ML/DL techniques \cite{BFL-CIDS}. Such scenario raises significant issues that need to be considered \cite{rahman2020internet}. First, the disclosure of IoT devices’ local data could represent a privacy concern for end users, since an attacker could even infer users’ daily habits by analyzing the traffic of their devices (e.g., wearables). This aspect could also pose an issue for a specific company where IoT devices share their network traffic with third parties. Second, given the dynamism of typical IoT environments, the time required to detect a potential attack could become a key aspect (or a limitation if the computing time is considerable) to prevent its spread in a certain network. In the case of using typical cloud data centers, the latency derived from the communication of a large quantity of data with data centers could be unaffordable or it could decrease the effectiveness of the IDS deployment. Although recent approaches propose the use of fog/edge computing \cite{eskandari2020passban} to balance the computing resources in the IDS implementation, this solution still raises privacy concerns as devices’ data is shared with external entities (i.e., fog/edge nodes). Third, many IoT scenarios are comprised of resource-constrained devices communicating through wireless technologies with limited bandwidth and throughput. Therefore, the constant communication of devices' network data could represent a high overhead for IoT networks with a high number of connected devices.

To address these issues, there is a need for decentralized approaches with on-device learning in which devices themselves could perform local processing on their own network traffic data. As described by \cite{rahman2020internet}, a distributed or \textit{self-learning} approach is a potential solution in which devices perform local training without interacting with each other. However, in this approach, devices are not able to improve their learning capacity based on the learning process of the other devices in the network. As an alternative, Federated Learning (FL) was proposed in 2016 \cite{mcmahan2017communication} as a collaborative learning approach in which devices still interact each other through a centralized entity without the need for sharing their data. Figure \ref{Fig:FLscheme} shows an overview of the centralized, distributed, and federated learning approaches. 

In a typical FL scenario, end devices do not share their data. Instead, they update the information onto the global model based on local calculations on their own data. These nodes are typically called \textit{clients} or \textit{parties}, and the entity responsible for aggregating such local updates is called \textit{coordinator} or \textit{aggregator}. The training process is divided into a set of \textit{rounds}, in which clients interact with the coordinator to update the global model until a certain number of rounds is performed or a certain accuracy is achieved. In particular, the main steps of each training round comprise \cite{mcmahan2017communication} \cite{rahman2020survey}:

\begin{enumerate}
    \item The coordinator selects a subset of clients. For this purpose, different aspects can be considered; for example, in an IoT scenario, devices' computation/communication resources can be used to select the most suitable clients to participate in the training round \cite{nishio2019client}.
    \item The coordinator sends the parameters/weights of the global model to the selected clients. 
    \item The different clients update the global model's parameters/weights through a training process by using Stochastic Gradient Descent (SGD) with their local data. In the case of an IDS system, the training is intended to be performed by using the local network traffic of each client.
    \item Then, the clients send their updated model's parameters/weights back to the coordinator. Depending on the aggregation algorithm being used, the coordinator aggregates all the parameters/weights to build a new global model, which will be used in the next training round. Although FedAvg is the most widely used aggregation algorithm \cite{mcmahan2017communication}, there is a plethora of alternative algorithms that can be considered for this process, such as FedProx \cite{li2018federated} or the recent Fed+ \cite{fed+paper}, which is used in our evaluation.
\end{enumerate}

The application of FL in IoT scenarios has attracted a huge interest in recent years due to its benefits compared to traditional centralized learning approaches. However, there are still significant challenges to be considered, such as communication and computing requirements or potential security and privacy attacks \cite{mothukuri2020survey, feraudo2020colearn}. In the context of IoT, the FL application for IDS is still in its infancy, and existing proposals are often based on unrealistic settings and data distributions. These efforts are described in the next section.
\section{Related Work}\label{sec:related}
As previously mentioned, the use of FL has attracted a significant interest in recent years due to its characteristics and strengths, which can be exploited in different IoT scenarios \cite{nguyen2021federated}. In this context, recent works have proposed the application of FL to improve IDS. To classify these works, we have considered various aspects, such as: analyzed attacks, training datasets, ML/DL algorithms to detect such attacks, aggregation methods, and implementation frameworks.  An overview of these proposals is shown in Table \ref{tab:taxonomy_table}.

\begin{table}[]
\tiny
\centering
\begin{tabular}{C{1.2cm}C{1.2cm}C{1.2cm}C{1.2cm}C{1.2cm}C{1.2cm}C{1.2cm}C{1.2cm}}

\textbf{Reference} & \textbf{Attack studied} & \textbf{Dataset} & \textbf{ML Model} & \textbf{FL implementation} & \textbf{Aggregation function} & \textbf{Training parties} & \textbf{Training rounds} \\ \hline \hline

 \cite{FLEAM} & 2 & Simulated traffic & GRU & - & FedAvg & - & - \\ \hline

\cite{DIOT} & 3 & Generated & GRU & - & FedAvg  & 14 & 3\\ \hline

\cite{BFL-CIDS} & 1, 4- 6 & KDDCup99 & MLP, DT, SVM, RF & - & FedAvg & - & 50\\ \hline

\cite{FLMimicIDS}  & 1, 4- 6 & NSL-KDD & MLPs & TensorFlow, Keras & FedAvg  & 10 & 20\\ \hline

\cite{rahman2020internet} & 1, 4-6 & NSL-KDD & NN & - & FedAvg & 4 & 1-5\\ \hline

\cite{FLHeteCohort} & 1, 7-11 & CSE-CIC-IDS2018 & NN & TensorFlow & FedAvg & 1-50 & 10000\\ \hline

\cite{BNNSIDFL} & 1, 7-10, 12 & CICIDS2017, ISCX Botnet 2014 & Binarized NN & TensorFlow & signSGD & - & -\\ \hline

\cite{CLCAIoT40} & 1, 4-6, 13- 20 & KDD, NSL-KDD, UNSW-NB15, N-BaIoT & Deep belief network & - & FedAvg & - & -  \\ \hline

\cite{LocKedge} & 1, 2, 12, 21, 28, 29 & BoT-IoT & NN & - & FedAvg & 4 & 1000\\ \hline 

\cite{rey2021federated} & 3, 20 & N-BaIoT & MLP, autoencoders & own library \cite{fediotguard} & FedAvg, Coordinate-wise median/trimmed mean & 8 & 1-29\\ \hline

\cite{DEEPFED} & 1, 17, 22 & \cite{morris2014industrial} & Convolutional NN, GRU & Flask \cite{grinberg2021flask}, Keras \cite{ketkar2017introduction} & FedAvg & 3-7 & 2-10 \\ \hline

\cite{mothukuri2021federated} & 23, 24 & Modbus dataset & GRU & Pytorch/PySyft & FedAvg & - & 1-40\\ \hline

Our approach & 1, 2, 15, 22, 24- 27 & CIC-ToN-IoT & Logistic regression & IBMFL & FedAvg, Fed+ & 4/10 & 1-300\\ \hline 

\end{tabular}
\caption{Classification of existing works on FL-enabled IDS for IoT. \newline Legend. 1: DoS, 2: DDoS, 3: Mirai, 4: U2R, 5: R2L, 6: Probe, 7: Web, 8: Bruteforce, 9: Infiltration, 10: Botnet, 11: DDOS+PortScan, 12: PortScan, 13: Fuzzers, 14: Analysis \cite{7348942}, 15: Backdoor, 16: Generic \cite{7348942}, 17: Reconnaissance, 18: Shell code, 19: Worm, 20: BASHLITE, 21: Keylogging, 22: Injection, 23: Flooding, 24: MITM, 25: XSS, 26: Password, 27: Scanning, 28: Data theft, 29: OS Fingerprinting}
\label{tab:taxonomy_table}
\end{table}

Based on our analysis, we note that some of the proposed works use their own generated or simulated dataset, 
For example, \cite{FLEAM} integrated an FL approach with fog computing, where fog nodes collaborated for detecting DDoS attacks. For this purpose, authors use Gated Recursive Units (GRUs) \cite{dey2017gate} as ML technique, and FedAvg as the aggregation algorithm. Also based on GRU, \cite{DIOT} proposes the creation of communication profiles associated to IoT devices that are used to detect potential attacks. In this case, the dataset is generated from real devices and the use of traffic associated with the Mirai botnet \cite{antonakakis2017understanding}. As these works are not based on publicly available datasets, it is difficult to assess the suitability of their proposed approach. Furthermore, in the case of \cite{FLEAM}, authors do not provide performance details, such as the different numbers of participating clients and training rounds.

While other FL-enabled IDS approaches have been proposed for IoT scenarios, they are not based on datasets with traffic associated with such devices. In this direction, \cite{BFL-CIDS} evaluates different ML models, such as decision trees, Support Vector Machines (SVM), Random Forest and MultiLayer Perceptron (MLP) in a federated environment in which the aggregation process is enabled through the use of blockchain. The proposed approach is based on intermediate nodes to perform local training using IoT devices’ data, as well as the KDDCup99 dataset \cite{stolfo1999kdd}. Moreover, \cite{FLMimicIDS} uses the NSL-KDD dataset \cite{tavallaee2009detailed} and MLP as the ML model for a FL-enabled IDS system. The approach is based on the concept of mimic learning in which a \textit{student} model is trained with a public dataset, which is labelled with a \textit{master} model trained with sensitive data. Also based on the NSL-KDD dataset, \cite{rahman2020internet} uses neural networks to propose a FL-enabled IDS considering three scenarios according to different data distributions regarding attack types. The use of neural networks is also proposed by \cite{FLHeteCohort}, which integrates a differential privacy approach \cite{wei2020federated}. For this purpose, authors consider a scenario with non-iid data using the CSE-CIC-IDS2018 dataset \cite{sharafaldin2018toward}. Moreover, \cite{BNNSIDFL} employs Binarized Neural Networks (BNNs) \cite{hubara2016binarized} in edge devices to reduce the overhead of traditional neural networks. The proposal is based on the datasets CICIDS2017 \cite{sharafaldin2018toward} and ISCX Botnet 2014 \cite{beigi2014towards}, as well as the aggregation algorithm signSGD \cite{bernstein2018signsgd} in order to reduce the overhead during the communication of model updates.

Besides previous works, recent efforts consider IoT-specific datasets to develop FL-enabled IDS in these scenarios. In particular, \cite{CLCAIoT40} proposes the use of deep belief networks \cite{hinton2009deep} to be deployed in IoT gateways to detect potential attacks on a certain IoT subnet. Then, the different models are aggregated through FL. The proposed approach uses several datasets, such as the N-BaIoT \cite{meidan2018n} dataset, which includes IoT devices’ traffic. However, authors do not provide information on the implementation being used or evaluation details considering aspects such as data distribution, number of clients or training rounds. This dataset is also used by \cite{rey2021federated}, which proposes a binary classification approach based on supervised learning (using MLP) and unsupervised learning (using autoencoders). Additionally, the proposed approach uses different aggregation methods based on \cite{yin2018byzantine}, which are compared considering different types of attack. In this case, it should be noted that authors created a balanced dataset with the same number of samples and proportion of classes for all devices. This distribution could be compared with our balanced scenario described in Section \ref{sec:partitioning}. Moreover, the Bot-IoT dataset \cite{koroniotis2019towards} is used by \cite{LocKedge}, which proposes multiclass classification based on neural networks together with Principal Component Analysis (PCA) in an edge-based network architecture with IoT gateways. The proposal distributes the dataset in four clients according to attackers’ IP address; however, details on the implementation being used and data distribution in the different parties are not described. Additionally, other works on the use of FL for IDS in IoT are based on specific datasets for industrial environments. In this direction, \cite{DEEPFED} integrates Convolutional Neural Networks (CNN) and GRUs for the detection of different attacks using the dataset described in \cite{morris2014industrial}. Furthermore, \cite{mothukuri2021federated} also uses GRU with a dataset based on the well-known Modbus protocol \cite{morris2014industrial}.

Our literature analysis demonstrates that the development of FL-enabled IDS approaches for IoT is still in its infancy. On the one hand, while most of the previous works are intended to be considered in such scenarios, they are not based on datasets with IoT devices' network traffic. On the other hand, we note that a significant amount of the previous works do not provide information about the implementation being used, or details related to the evaluation process, such as number of clients or training rounds. Furthermore, most of the works do not describe the data distribution among the different clients, or they consider scenarios where clients’ data are associated to a portion of the dataset that includes the same number of samples for each attack being considered. However, as discussed in previous works \cite{zhao2018federated}, the performance of FL can be reduced in the case of scenarios with non-iid and highly skewed data. While these aspects have not been evaluated in the context of FL-enabled IDS, our work provides an exhaustive evaluation under different data distributions using the recently proposed ToN\_IoT \cite{ToNIoTPaper} dataset, which includes several IoT-related attacks. To cope with the impact of non-iid data, we compare the performance of the typical FedAvg algorithm with a recent approach called Fed+ \cite{fed+paper}. To the best of our knowledge, this is the first approach evaluating the impact of non-iid data on the development of FL-enabled IDS for IoT.
\section{Methodology}\label{sec:methodology}

Before describing our evaluation results for the proposed FL-enabled IDS for IoT considering non-iid data, in this section we explain the main processes and assets used for this purpose. They include the dataset selection, data distribution among several FL clients, as well as the classifier technique and aggregation functions being considered.

\subsection{Dataset selection}\label{sec:dataset_selection}
For the development of our FL-enabled IDS proposal for IoT, a key aspect is the selection of an appropriate dataset. As described in the previous section, recent approaches are based on obsolete and generic network traffic datasets, which do not consider IoT-specific protocols and attacks. Furthermore, as described by \cite{rey2021federated}, most of the datasets for IDS were not conceived to be used in an FL environment, as they cannot be properly distributed among different clients. Therefore, our analysis is focused on IoT datasets for IDS that can be divided by IP address or device \cite{rey2021federated}, namely Bot-IoT \cite{koroniotis2019towards}, N-BaIoT \cite{meidan2018n}, MedBIoT \cite{guerra2020medbiot}, IoTID20 \cite{ullah2020scheme} and ToN\_IoT \cite{ToNIoTPaper}. In the case of ToN\_IoT, we consider the CIC-ToN-IoT dataset \cite{cictoniot}, which is generated from the original pcap files of ToN\_IoT. An overview of these datasets is shown in Table \ref{table:comparison1}, in which they are compared according to several aspects, such as number of features and samples, attacks, the use of labelled data, or their testbed. 

\begin{table}[]
\tiny
\centering
\begin{tabular}{C{1.2cm}C{1.3cm}C{1.2cm}C{1.2cm}C{1.3cm}C{3.5cm}C{0.8cm}C{0.8cm}C{0.8cm}}
\textbf{Dataset}& \textbf{Training/ testing sets?} & \textbf{\# features} & \textbf{\# samples} & \textbf{Normal/malign flow ratio} & \textbf{Attacks} & \textbf{Data labelled?} & \textbf{Best-features set} & \textbf{Realistic testbed?}\\
 \hline  \hline
 Bot-IoT \cite{koroniotis2019towards} & Y & 46 & 73,370,443 & 0.00013:1 & \vspace{-1.3em}\flushleft{PortScan, OS Fingerprinting, DoS/DDoS, Data Theft, Keylogging} & Y & Y & Y\\
 \hline
 N-BaIoT \cite{meidan2018n} & N  & 115 & 7,062,606 & 0.07:1 & \vspace{-1.3em}\flushleft{Mirai Bot, BashLite Bot} & Y & N & Y\\
 \hline
 MedBIoT \cite{guerra2020medbiot} & N & 100 & 17,845,567 & 2.36:1 & \vspace{-1.3em}\flushleft{Mirai Bot, BashLite Bot, Torii Bot} & Y & N & Y\\
 \hline
     IoTID20 \cite{ullah2020scheme}  & N & 83 & 625,784 & 0.06:1 & \vspace{-1.3em}\flushleft{Mirai Bot, MITM, PortScan, OS Fingerprinting} & Y & Y & Y\\
 \hline
 CIC-ToN-IoT \cite{cictoniot} & N & 83 & 5,351,760 & 0.88:1 & \vspace{-1.3em}\flushleft{Backdoor, DoS, DDoS, Injection, MITM, Password, Ransomware, Scanning, XSS} & Y & N & Y\\
 \hline

\end{tabular}
\caption{Comparison between relevant contemporary intrusion datasets for IoT (N=NO, Y=YES)\label{table:comparison1}}
\end{table}


A common aspect of the different datasets is that they are based on realistic testbeds, as well as labelled data considering different types of attack. Bot-IoT is the only analyzed dataset that provides training and testing sets. Furthermore, this dataset and IoTID20 identify a set of best features to be considered. However, we note that most of the datasets suffer from a significant imbalance between benign and attack traffic that can negatively affect the ML/DL approach, so that oversampling/undersampling could be required. In this direction, we note that the ToN\_IoT dataset provides the best ratio between benign and attack traffic. Furthermore, this dataset considers a broader diversity of attack types compared to the other datasets being analyzed. For example, N-BaIoT and MedBIoT focus on particular attacks that are launched by IoT devices composing a botnet. However, they do not consider other attacks, such as DDoS/DoS or MITM that should be considered in IoT environments.

Moreover, while the different datasets are based on realistic testbeds, ToN\_IoT is built using an IoT/IIoT testbed composed by edge/fog nodes and cloud components to simulate an IoT/IIoT production environment. Furthermore, ToN\_IoT is the only dataset that considers data from sensor readings and telemetry data, which can be used to detect additional attacks (beyond the network level) in such environments. Although ToN\_IoT has been used in recent works (e.g., \cite {ferrag2021deep}), to the best of our knowledge, this is the first effort to consider ToN\_IoT in a FL setting. 

\subsection{ToN\_IoT partitioning}\label{sec:partitioning}
To create the three proposed scenarios based on different data distributions, we use the CIC-ToN-IoT dataset \cite{cictoniot}, which was generated through the CICFlowMeter tool \cite{lashkari2017characterization} from the original pcap files of the ToN-IoT dataset, as previously described. Such tool was used to extract 83 features, which were reduced by removing those with a non-numeric value (e.g., flow ID). Then, we separate the samples of the whole dataset according to the destination IP address, and select the 10 IP addresses with more samples. Those observations constitute our dataset. Such resulting dataset contains 4.404.084 samples, which represent 82,29\% of the original CIC-ToN-IoT.

From this dataset, we create three scenarios to evaluate the impact of different data distributions on the performance of our multiclass classifier to detect attacks. The datasets of such scenarios are available at \cite{datasets}. Specifically, we use Shannon entropy \cite{bonachela2008entropy} to measure the imbalance of the different local datasets of each FL client. In particular, given a dataset of length $n$, and $k$ classes of size $c_i$, the balance between the classes is given by the formula:$$\text{Entropy}=\frac{-\Sigma_{i=1}^k\frac{c_i}{n}\log\frac{c_i}{n}}{\log k} $$where the function is equal to 0 if all classes are 0 except one, and is equal to 1 if all $c_i=\frac{n}{k}$. 
Furthermore, it should be noted that we consider that each FL client is represented by a single IP address. In this context, $n$ is the number of network flows, $k$ is the number of the attack classes and $c_i$ is their size.

\subsubsection{Basic scenario}\label{sec:basic}
In this scenario, each FL client’s dataset is based on the network traffic of the corresponding IoT device. As described in Table \ref{tab:description_total}, in this case the distribution of classes and samples among the different nodes is highly unbalanced. Indeed, party 7 only has benign traffic samples, while parties 1 and 3 only have 2 samples of XSS attack. Consequently, these parties have the lowest Shannon entropy value. This scenario represents a typical situation in a certain IoT network in which specific devices can be victims of several attacks while other devices perform their intended operation and they are not subject to attacks. However, as described in Section \ref{sec:evaluation}, the straightforward application of FL in this scenario could result in poor performance and convergence issues.

\subsubsection{Balanced scenario}\label{sec:balanced}
In this case, we select a portion of our dataset, which is distributed among the 10 parties, so that each party has the same number of samples of each class. Therefore, as shown in Table \ref{tab:description_total}, all the parties have the same Shannon entropy value. As will be described in Section \ref{sec:evaluation}, such balanced scenario presents better performance; however, in this case, each FL client could have samples of other nodes, so that it can result in privacy issues depending on the scenario being considered. It should be noted that such scenario can be compared with similar settings in previous works, such as \cite{rey2021federated}, which uses a version of the N-BaIoT dataset where the number of samples and the class proportions are the same for all devices. 

\subsubsection{Mixed scenario}\label{sec:mixed}
The mixed scenario is generated to achieve a tradeoff between the two previous settings in which each party maintains its own samples, but they are locally balanced. In particular, we select the parties with a Shannon entropy value higher than a certain threshold (0.2), that is, parties 0, 2, 4 and 5. After this initial filtering step (due to the fact that the parties' classes are not well balanced) we use a simple instance selection mechanism that removes some of the samples from the predominant classes until we reach the Shannon entropy within a range of values. Having this set in between 0.66 and 0.71, we obtain a dataset that represents a compromise between the basic scenario where no balancing was used, and the balanced scenario where we artificially distributed the dataset among the 10 parties.

\begin{table*}[]
\tiny
\centering
\begin{tabular}{lC{0.8cm}C{0.9cm}C{0.8cm}C{0.8cm}C{0.8cm}C{0.8cm}C{0.8cm}C{0.8cm}C{0.8cm}C{0.8cm}C{0.8cm}>{\columncolor[gray]{0.8}}c}
\textbf{Scenario} & \textbf{Party} & \textbf{Total samples} & \textbf{Benign} & \textbf{XSS} & \textbf{Injection} & \textbf{Password} & \textbf{Scanning} & \textbf{MITM} & \textbf{DDoS} & \textbf{Dos} & \textbf{Backdoor} & \textbf{Entropy}\\ \hline \hline
 & 0 & \textbf{811504} & 42527 & 474520 & 140519 & 140519 & 13419 & - & - & - & - & 0.52041  \\ \cline{2-13} 
 
 & 1 & \textbf{763518}& 763516 & 2 & - & - & - & - & - & - & - & 0 \\ \cline{2-13} 
 
 & 2 & \textbf{740117} & 116540 & 594627 & 16271 & 1138 & 10923 & 253 & 202 & 145 & 18 & 0.28669\\ \cline{2-13} 
 
 & 3 & \textbf{519806} & 519804 & 2 & - & - & - & - & - & - & - & 0 \\ \cline{2-13} 

Basic & 4 & \textbf{424531} & 2794 & 307962 & 66812 & 38009 & 8954 & - & - & - & - & 0.38890 \\ \cline{2-13} 

 & 5 & \textbf{330956} & 10537 & 206036 & 44043 & 67431 & 2909 & - & - & - & - & 0.47291 \\ \cline{2-13} 

 & 6 & \textbf{223092} & 3587 & 209637 & 9868 & - & - & - & - & - & - & 0.11976 \\ \cline{2-13} 

 & 7 & \textbf{217737} & 217737 & - & - & - & - & - & - & - & - & 0.0002\\ \cline{2-13} 
 
 & 8 & \textbf{186891} & 8981 & 177910 & - & - & - & - & - & - & - & 0.08794\\ \cline{2-13} 
 
 & 9 & \textbf{185932} & 8551 & 177381 & - & - & - & - & - & - & - & 0.08511\\ \hline
 
Balanced & 0-9 & \textbf{43549} & 10000 & 10000 & 10000 & 10000 & 3500 & 20 & 18 & 10 & 1 & 0.7611\\ \hline

& 0 & \textbf{205946} & 42527 & 50000 & 50000 & 50000 & 13419 & - & - & - & - & 0.69858 \\ \cline{2-13}

Mixed& 2 & \textbf{42679}  & 10000 & 10000 & 10000 & 1138 & 10923 & 253 & 202 & 145 & 18 & 0.70266 \\ \cline{2-13}

& 4 & \textbf{71748} & 2794 & 20000 & 20000 & 20000 & 8954 & - & - & - & - & 0.66218 \\ \cline{2-13}
& 5 & \textbf{73446} & 10537 & 20000 & 20000 & 20000 & 2909 & - & - & - & - & 0.66888 \\ \hline
\end{tabular}
\caption{Description of the basic, balanced and mixed scenarios}
\label{tab:description_total}
\end{table*}

\subsection{Multiclass classification}
Considering the already described scenarios, we use a multiclass probabilistic classification model to classify the instances into benign or a specific type of attack. For this purpose, we apply the multinomial logistic regression \cite{bohning1992multinomial}, also called soft-max regression, due to its easy implementation and training efficiency. It can also interpret model coefficients as indicators of feature importance. Multinomial logistic regression is a simple extension of binary logistic regression \cite{LogReg} that allows for more than two categories of the dependent or outcome variable which do not present an order. As with most classifiers, the input variables need to be independent for the correct use of the algorithm.
Given the input $x$, the objective is to know the probability of $y$ (the label) in each potential class $p(y=c|x)$. The softmax function takes a vector $z$ of $k$ arbitrary values and maps them to a probability distribution as follows \[ \mbox{softmax}(z_i) = \frac{\exp(z_i)}{\sum_{j=1}^{k}\exp(z_j)}. \] In our case, the input to the softmax will be the dot product between a weight vector $w$ and the input vector $x$ plus a bias for each of the k classes: \[ p(y=c|x) = \frac{\exp(w_c \dot x + b_c)}{\sum_{j=1}^{k}\exp(w_j \dot x + b_j)}. \] The loss function for multinomial logistic regression generalizes the loss function for binary logistic regression and is known as the cross-entropy loss or log loss.

It should be noted that unlike previous works based on binary classifiers (e.g., \cite{rey2021federated}), we consider the detection of a specific attack as a key factor to dynamically deploy the most effective countermeasures to mitigate such threat. Furthermore, while other classifiers could be employed (and it represents part of our future work), our evaluation results are focused on the impact of different data distributions and non-iid data in the classifier performance.

\subsection{Aggregation functions}
As described in Section \ref{sec:background}, the local updates generated by each client in FL are combined through an aggregation function in each training round. The most basic aggregation function is represented by FedAvg \cite{mcmahan2017communication}, which generates the global model based on the average of the weights generated by the FL clients. In particular, let $W=(w_i)$ be the weights of the general model and $W^k=(w_i^k)$ the weights of the party $k$, then:
$$w_i = \sum \frac{d_i}{D}w^k_i,$$
where $D$ and $d_i$ are the total data size and data size of each party respectively.

However, as described in recent works \cite{HeteInFedAvg, li2019convergence, zhao2018federated}, the performance of FedAvg may be degraded in scenarios with non-IID and highly skewed data. While recent works propose alternative aggregation functions considering convergence and privacy aspects \cite{mothukuri2020survey}, in this work we consider a recent approach called Fed+ \cite{fed+paper}, which unifies several functions to cope with scenarios composed by heterogeneous data distributions. For this purpose, Fed+ relaxes the requirement of forcing all parties to converge on a single model (as in the case of FedAvg). In particular, let be the main objective in FedAvg:
$$\text{min} \, F(x) \, = \, \frac{1}{D}\sum f_i(x),$$
where $f_i$ is the local loss function of the party $i$. In the case of Fed+, the main objective is:
$$\text{min} \, F(x) \, = \, \frac{1}{D}\sum f_i(x) + \alpha_i B(x,C(X)),$$
where $B(\cdot,\cdot)$ is a distance function, and $C$ is an aggregate function that computes a central point of $x$. 

It should be noted that this work represents the first effort to use Fed+ to evaluate its impact in the context of FL-enabled IDS for IoT. As will be described in Section \ref{sec:evaluation}, the use of such approach mitigates the convergence issues of FedAvg specially in settings with non-iid and skewed data.

\section{Evaluation results}\label{sec:evaluation}
Based on the different aspects of the proposed methodology, in this section we describe our evaluation results. For this purpose, we consider the following metrics:
\begin{itemize}
    \item Accuracy: $\frac{TP+TN}{TP+FP+FN+TN}$
    \item Precision: $\frac{TP}{TP+FP}$
    \item Recall: $\frac{TP}{TP+FN}$
    \item F1-score: $2*\frac{Recall * Precision}{Recall + Precision}$
    \item False Positive Rate (FPR): $\frac{FP}{FP+TN}$
\end{itemize}
where TP: true positives, TN: true negatives, FP: false positives, and FN: false negatives. 

Precision, recall, F1-score, and FPR metrics are calculated for each scenario described in Section \ref{sec:partitioning}. In the case of multiclass classification, such metrics can be calculated by using micro, macro, and weighted averaging. The micro-averaging calculates the metrics using the total amount of TP, TN, FP, and FN, independently of the number of classes. The macro-averaging calculates each metric for each class independently, and then it uses the average of all the classes' values. 
Then, the weighted-averaging follows a similar approach to the macro-averaging, but instead of using the normal averaging, the average is weighted depending on the class size. As some of our scenarios are based on imbalanced datasets (see Section \ref{sec:partitioning}), we use the weighted-averaging for our evaluation.

Moreover, we train the model across 300 rounds for each scenario by considering one epoch for each training round. The number of epochs is a hyperparameter that defines the number of times that the learning algorithm will work through the entire training dataset in each specific client. One epoch means that each sample in the training dataset has updated the internal model parameters only once. Furthermore, the logistic regression algorithm is implemented by using scikit-learn SGDClassifier (Stochastic Gradient Descent). In particular, we choose a logarithmic loss function to use the logistic regression, and the norm $L_2$ in order to shrink model parameters towards the zero vector. Before the application of the ML/DL, the data is normalized. Furthermore, a ratio of 80-20 was defined between training and testing sets.

For our evaluation, we consider FedAvg and Fed+ as aggregation functions in our FL-enabled IDS approach. Furthermore, we also measure the accuracy of each client in a distributed scenario, where each party trains the model using their own data independently from the other parties (see Section \ref{sec:background}). It should be noted that we do not consider a centralised setting (in which devices send their data for training a model) because in that case all the classes would be represented in the dataset. Therefore, it would be unfair to compare such setting with a distributed/federated scenario in which clients only have traffic associated to their IP address, and only some of the classes are represented in their partial datasets. Nevertheless, for the sake of completeness, we measure the accuracy of the centralised setting and obtain a value of 0.724 using multinomial logistic regression. This value is close to 0.77, which represents the highest accuracy value obtained in the work describing the ToN\_IoT dataset \cite{ToNIoTPaper}.

\subsection{Basic scenario}
\begin{figure}[!htb]
    \centering
    \includegraphics[width=0.95\textwidth]{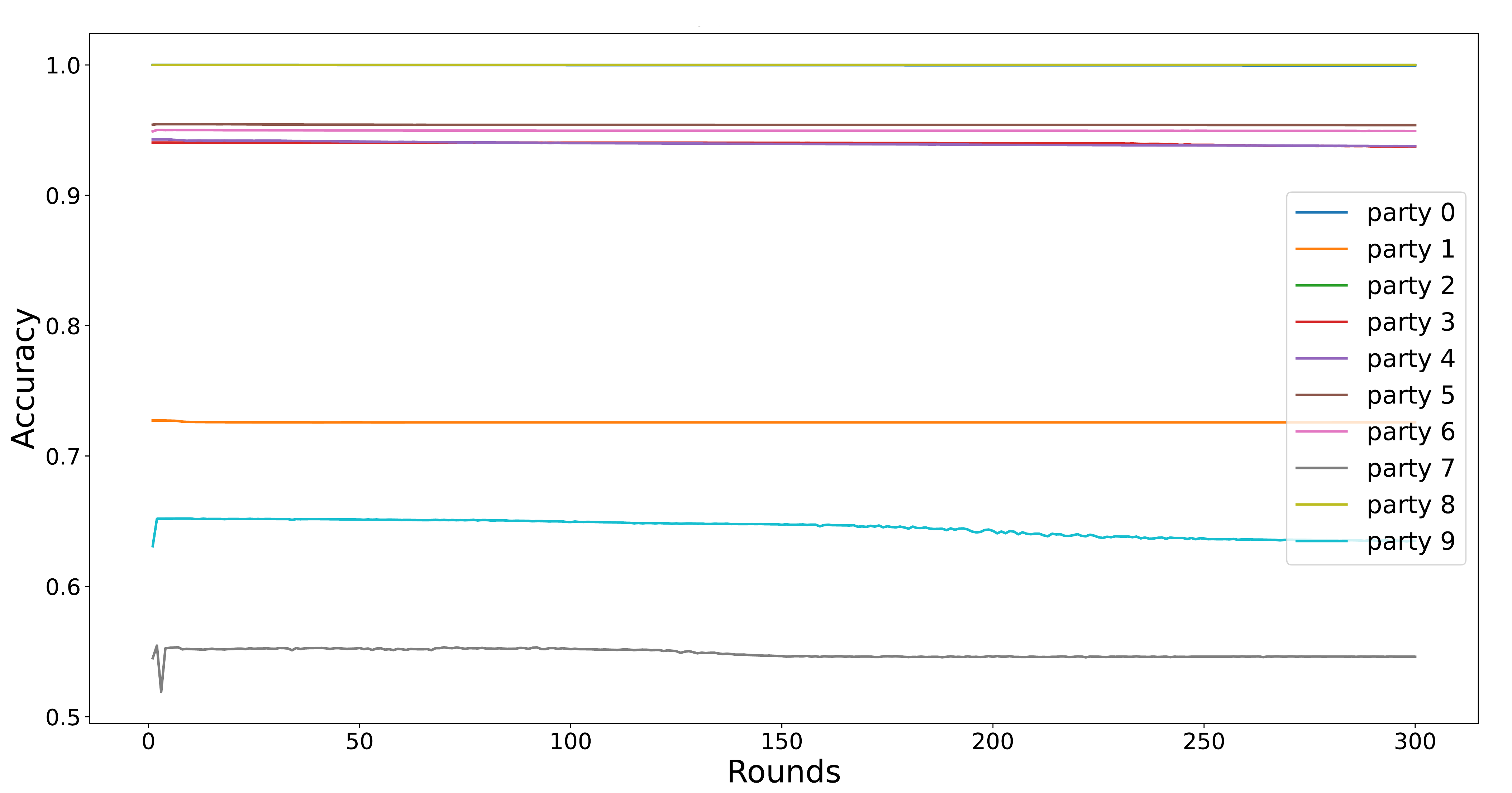}
    \caption{Basic scenario's accuracy with FedAvg}
    \label{fig:Acc_fa_real}
\end{figure}

\begin{figure}[!htb]
    \centering
    \includegraphics[width=0.95\textwidth]{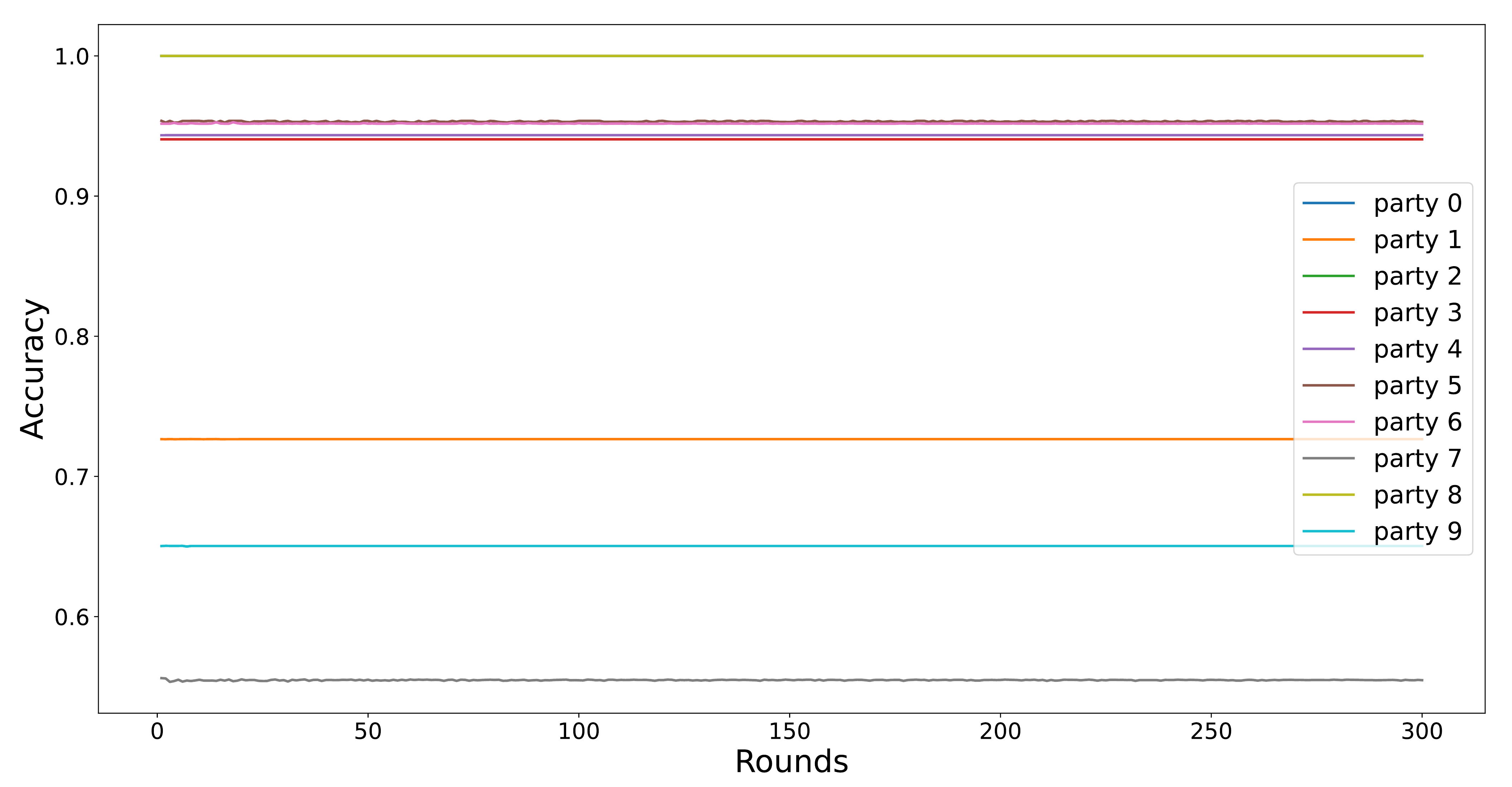}
    \caption{Basic scenario's accuracy with Fed+}
    \label{fig:Acc_fp_real}
\end{figure}

As described in Section \ref{sec:partitioning}, in this scenario, each party has the data corresponding to the traffic associated to a single IP address. Such scenario is characterized by a non-iid and highly skewed  data distribution. This aspect is reflected in Figure \ref{fig:Acc_fa_real} and Figure \ref{fig:Acc_fp_real}, which show the accuracy evolution of each client by using FedAvg and Fed+ methods, respectively. As shown, the accuracy value of each party remains stable throughout the training rounds. While the accuracy value seems high for parties 0, 2, 3, 4, 5, 6, and 8, this circumstance may be related to the heavily imbalanced dataset where accuracy may not be an exhaustive indicator because of the predominance of the data of the larger class (e.g., the legitimate traffic in this case). Then, accuracy is not fully representative since if a class represents the vast majority of the dataset, the classification process will provide a high accuracy even if only a single class is actually learned. However, the application of such model in a more balanced dataset may result in lower accuracy.
It should be noted that, according to Figure \ref{fig:Acc_fa_real}, the accuracy of parties 3, 4, 7, 9 is decreased after around 200 training rounds. This aspect could be related to the use of FedAvg as aggregation function that could represent convergence issues, as described by recent works \cite{li2019convergence}.

\begin{table}[]
\tiny
\centering
\begin{tabular}{ccc}
 & \textbf{Accuracy distributed} & \textbf{Accuracy federated} \\ \hline \hline
\textbf{Party 0} & 0.5526 & 1.0 \\ \hline
\textbf{Party 1} & 1 & 0.7293 \\ \hline
\textbf{Party 2} & 0.9435 & 1.0 \\ \hline
\textbf{Party 3} & 1 & 0.9402 \\ \hline
\textbf{Party 4} & 0.7283 & 0.9434 \\ \hline
\textbf{Party 5} & 0.6525 & 0.9525 \\ \hline
\textbf{Party 6} & 0.9412 & 0.9513 \\ \hline
\textbf{Party 7} & 1 & 0.5566 \\ \hline
\textbf{Party 8} & 0.9493 & 1.0 \\ \hline
\textbf{Party 9} & 0.9508 & 0.6527 \\  \hline
\end{tabular}
\caption{Comparison between distributed method and federated method.}
\label{tab:comparacionDisVsFed}
\end{table}

Table \ref{tab:comparacionDisVsFed} shows the accuracy of each party by considering the distributed and the federated scenario (using Fed+). It should be noted that parties with a low entropy (see Section \ref{sec:partitioning}) provide a higher accuracy in the distributed setting than in the federated scenario. This can be justified since parties with fewer classes and lower balance will classify better the samples of such predominant classes. Then, in the case of a federated environment, the weights of those parties with a few classes will be negatively influenced by the weights of other parties with more classes, because these parties detect different and additional types of attacks.

\begin{figure*}[!htb]
    \centering
    \includegraphics[width=\textwidth]{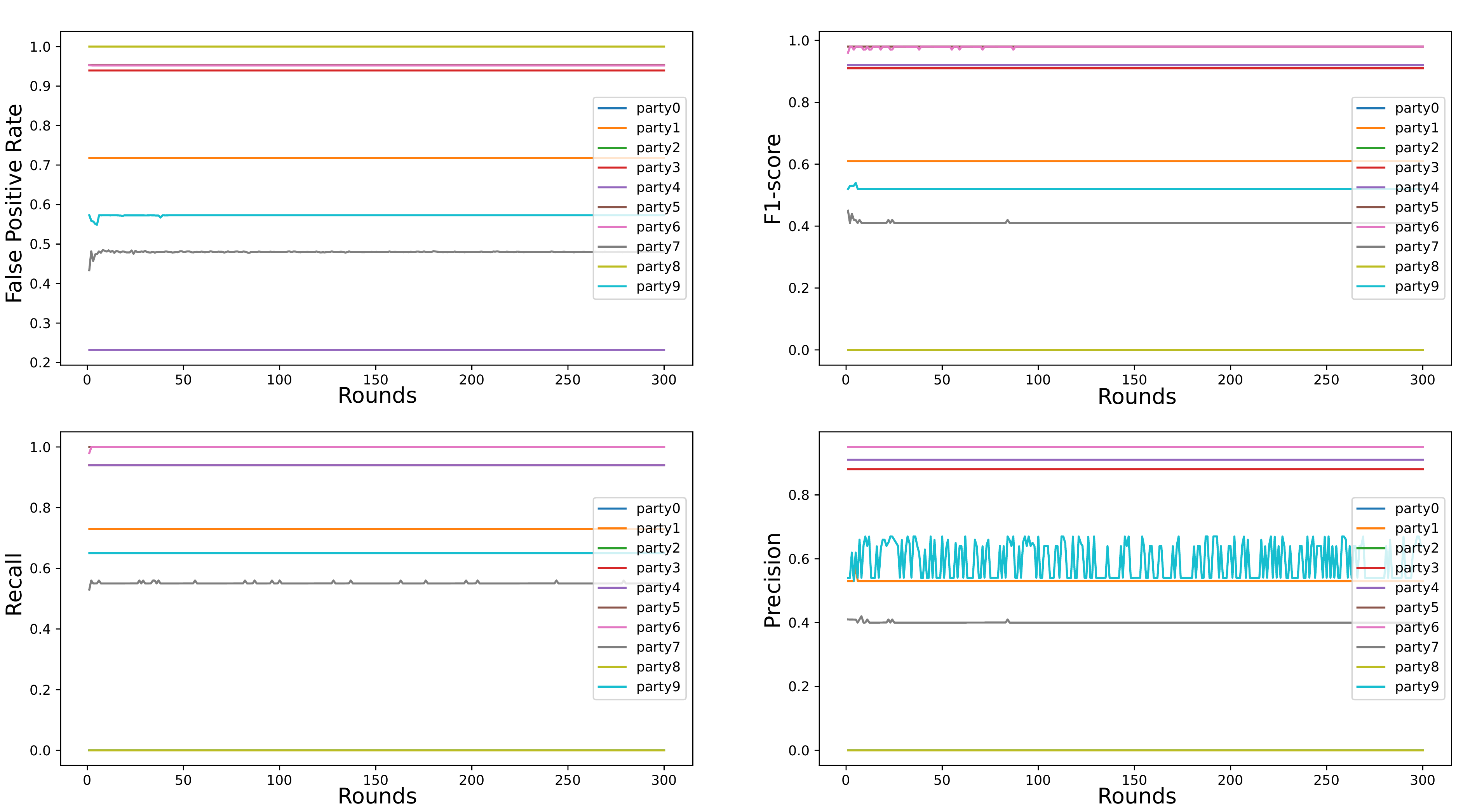}
    \caption{Basic scenario's FPR, F1-score, recall and precision with Fed+}
    \label{fig:Metrics_fp_real}
\end{figure*}

As shown in Figure \ref{fig:Metrics_fp_real}, the other metrics (beyond accuracy), calculated with Fed+, remain stable through the rounds, following a similar trend as the accuracy. The parties with a high FPR have poor results in terms of the others metrics. The values in recall, F1-score and precision of these parties are similar to the ones in the accuracy, except for party 2 and party 8, which provide 0 for precision and recall (and consequently in F1-score), and 1 for FPR. This situation can arise in scenarios with unbalanced datasets (like in this case), where a high accuracy is obtained (due to a high TN ratio) but recall and precision remain low (because of a low value for TP ratio)

Previous results demonstrate that the direct application of FL to scenarios with non-iid and highly skewed data could lead to undesirable results. Therefore, there is a need to consider a suitable client/instance selection process to make the dataset more balanced among the different clients in terms of number of classes and samples. The evaluation results for the balanced and mixed scenarios demonstrate the importance of such process, and are described below. 


\subsection{Balanced scenario}
\begin{figure}[!htb]
    \centering
    \includegraphics[width=0.95\textwidth]{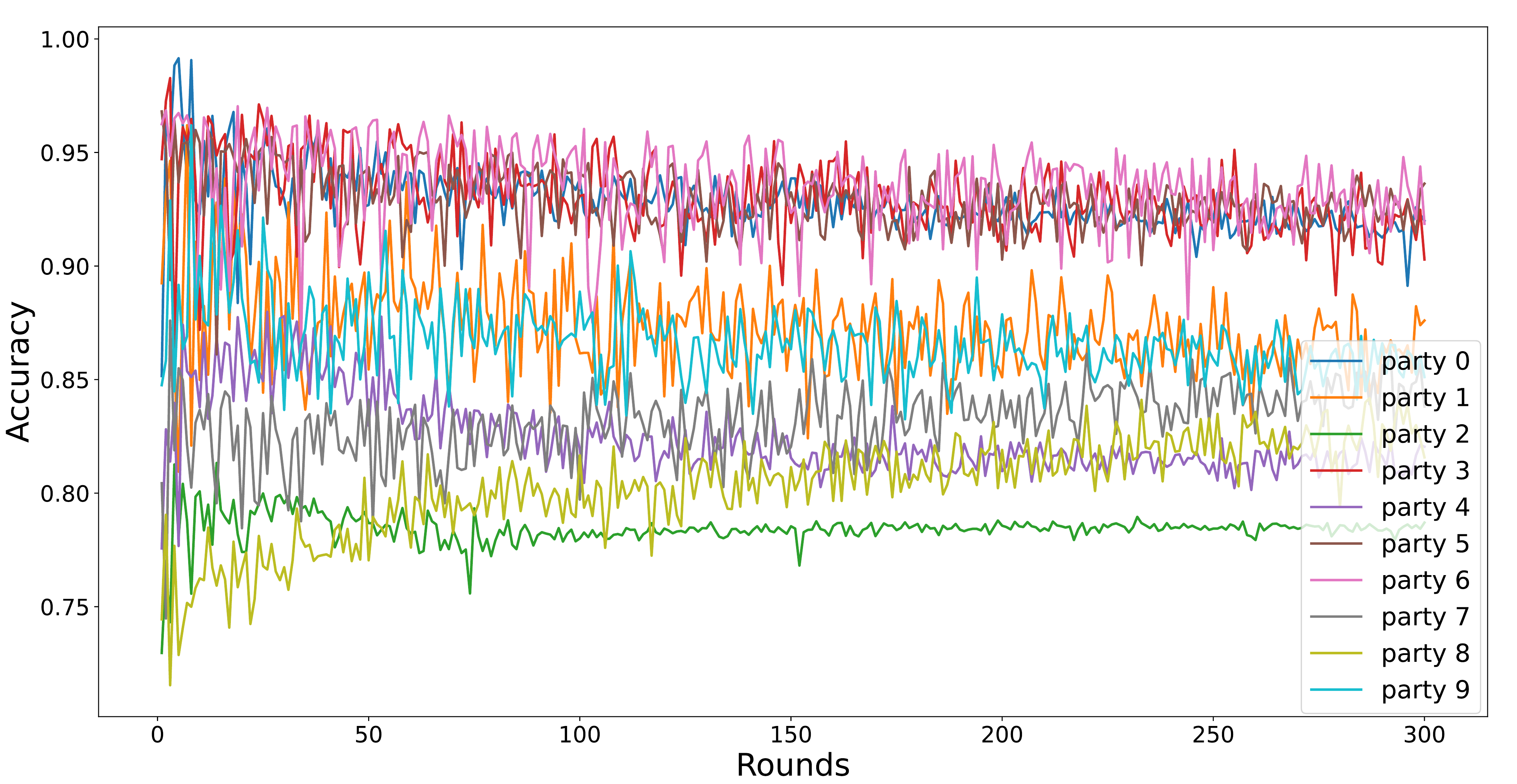}
    \caption{Balanced scenario's accuracy with FedAvg.}
    \label{fig:Acc_fa_ideal}
\end{figure}

\begin{figure}[!htb]
    \centering
    \includegraphics[width=0.95\textwidth]{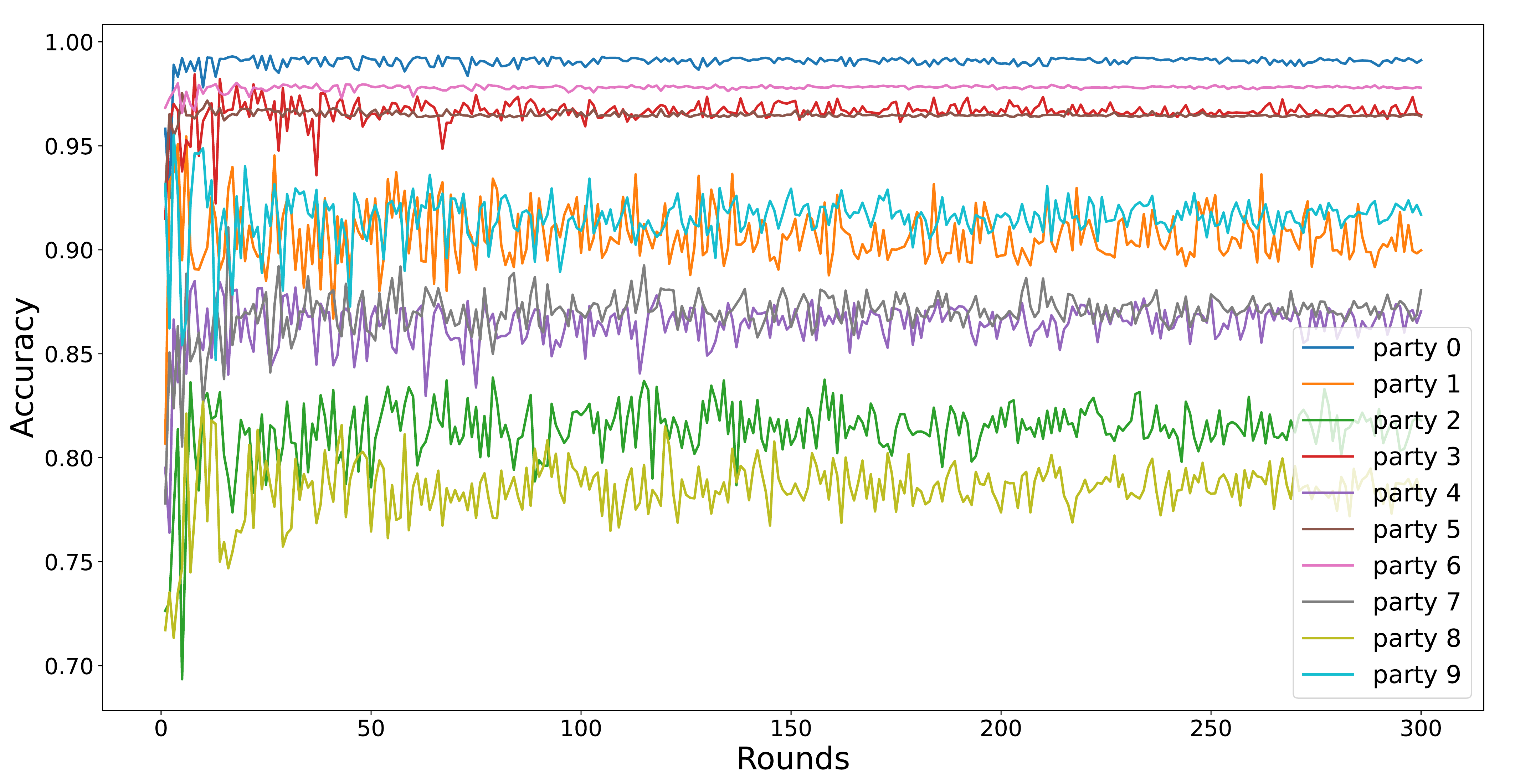}
    \caption{Balanced scenario's accuracy with Fed+.}
    \label{fig:Acc_fp_ideal}
\end{figure}

In this scenario, the data is equally distributed among parties according to the description provided in Section \ref{sec:balanced}. Figure \ref{fig:Acc_fa_ideal} and Figure \ref{fig:Acc_fp_ideal} show the evolution of the parties' accuracy by using FedAvg and Fed+ algorithms respectively. In the case of FedAvg, parties with a high accuracy obtain a decrease of such value throughout the rounds. For parties with a low accuracy, the evolution is similar to the Fed+ case. Furthermore, as shown in Figure \ref{fig:Acc_fp_ideal}, there is a clear increment in the accuracy for all parties that remain stable (between around 0.8 and 1) after about 50 rounds. 

\begin{figure*}[!htb]
    \centering
    \includegraphics[width=0.95\textwidth]{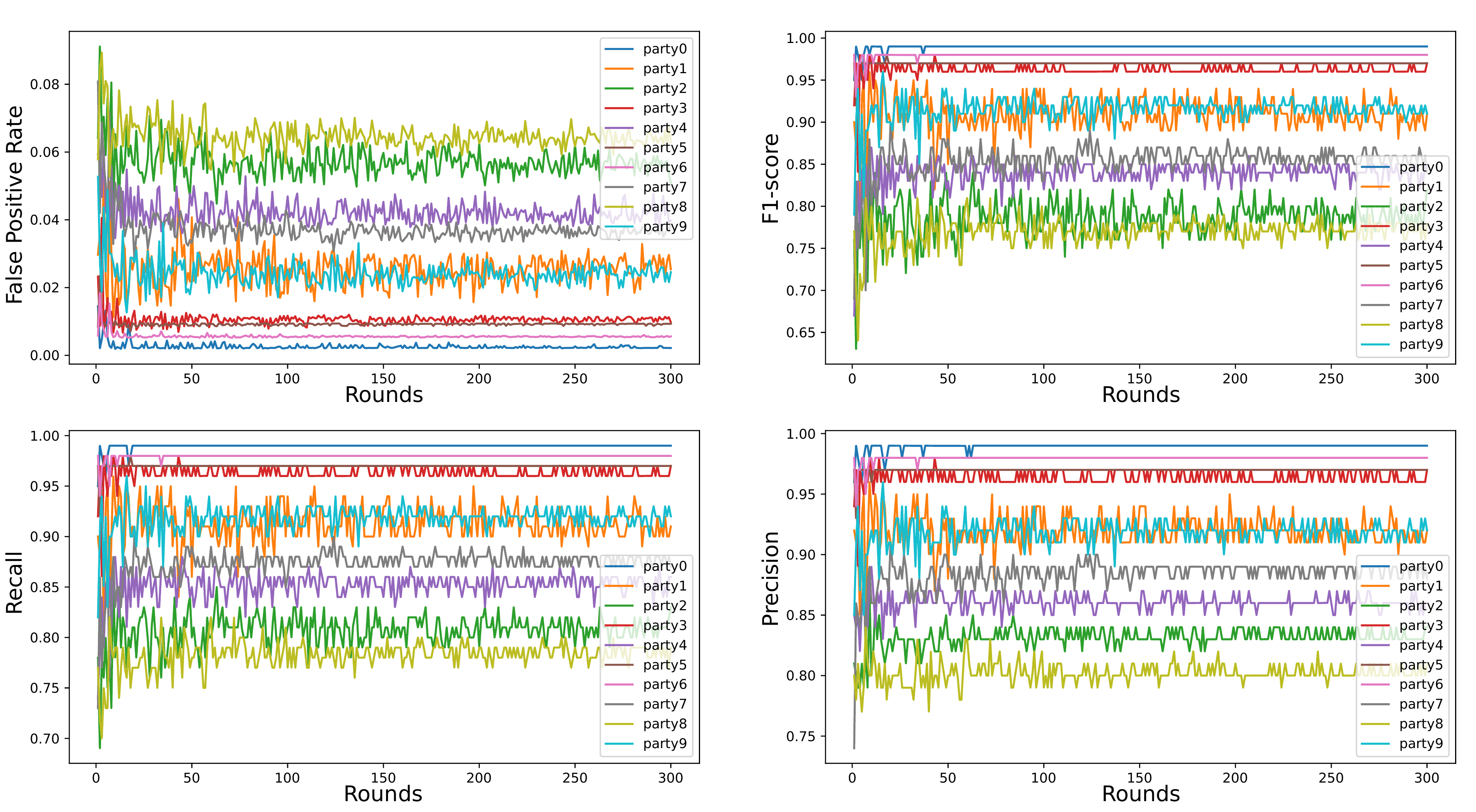}
    \caption{Better balanced scenario's precision, recall, F-1 score and FPR with fed+.}
    \label{fig:Metrics_fp_ideal}
\end{figure*}

Furthermore, the evolution of FPR, F1-score, recall and precision metrics in the case of Fed+ are shown in Figure \ref{fig:Metrics_fp_ideal}. In particular, the value of recall, F1-score and precision increase throughout the rounds with a similar trend as the accuracy. Moreover, the FPR value decreases throughout the rounds until it converges to a lower value. Compared with the results for the basic scenario, these metrics have values akin to the accuracy following a similar trend.

According to the obtained results, this scenario shows a better evolution in the parties for the different metrics being considered compared to the basic scenario. In particular, in the case of Fed+, all the parties improve such metrics throughout the initial 50 rounds, when their values remain stable. However, in the case of FedAvg, these value drop for some of the parties. Therefore, spite this scenario was artificially balanced, so that the parties have samples all the different attacks, the use of FedAvg still could lead to convergence issues. This could be due to the fact that even with a more balanced dataset among the different parties, the number of samples of each attach in every party still remains unbalanced. 


\subsection{Mixed scenario}
\begin{figure}[!htb]
    \centering
    \includegraphics[width=0.95\textwidth]{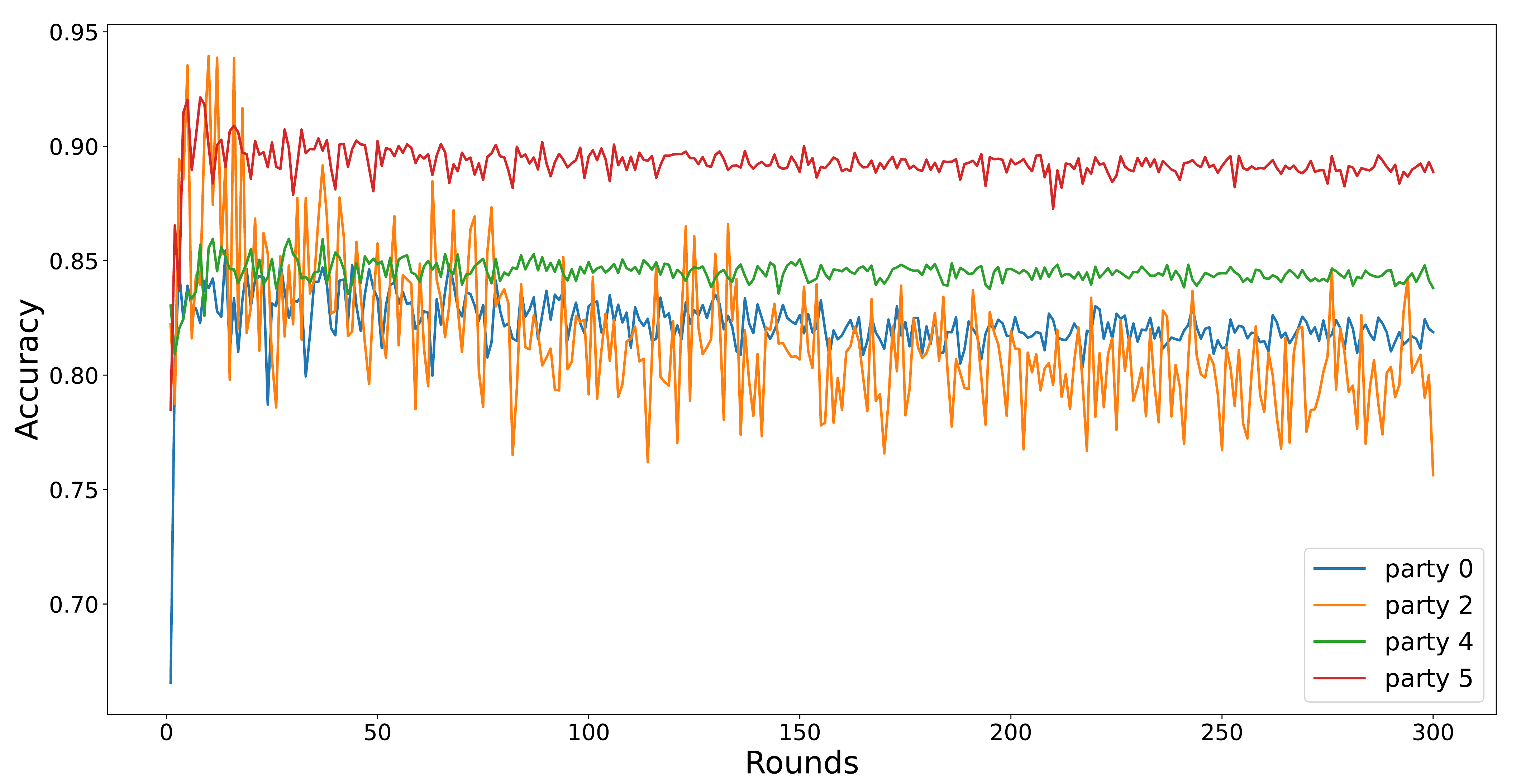}
    \caption{Mixed scenario's accuracy with FedAvg.}
    \label{fig:Acc_fa_mix}
\end{figure}

\begin{figure}[!htb]
    \centering
    \includegraphics[width=0.95\textwidth]{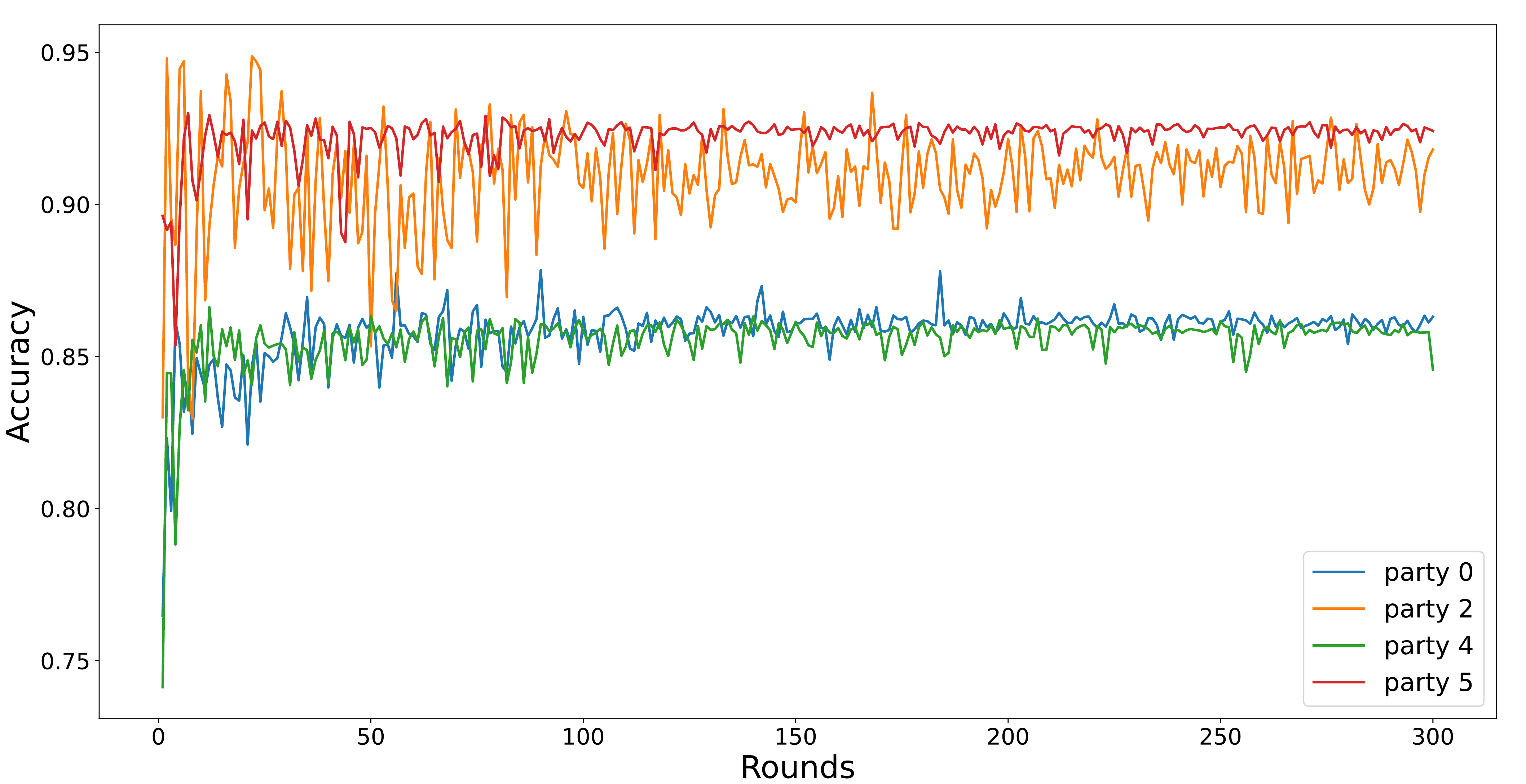}
    \caption{Mixed scenario's accuracy with Fed+.}
    \label{fig:Acc_fp_mix}
\end{figure}

The data distribution for this scenario is described in Section \ref{sec:mixed}. Figure \ref{fig:Acc_fa_mix} shows the accuracy evolution for each party when FedAvg is used. According to it,  there is a clear decrease in the accuracy of party 2 until about round 200, and such trend is also observed for party 0 after a significant increase in the very initial rounds. In the case of party 4 and party 5, the accuracy value remains stable. The decrease of accuracy is due to the unbalance of the scenario in which parties 0, 4 and 5 only have a subset of attack types. 
Then, Figure \ref{fig:Acc_fp_mix} shows the accuracy evolution of the different parties with Fed+, in which accuracy values grow until a certain number of rounds (about 50) when they remain stable. In the case of party 2, accuracy is more oscillating due to the fact that such party has samples of all the different classes in its local dataset. 


\begin{figure*}[!htb]
    \centering
    \includegraphics[width=0.95\textwidth]{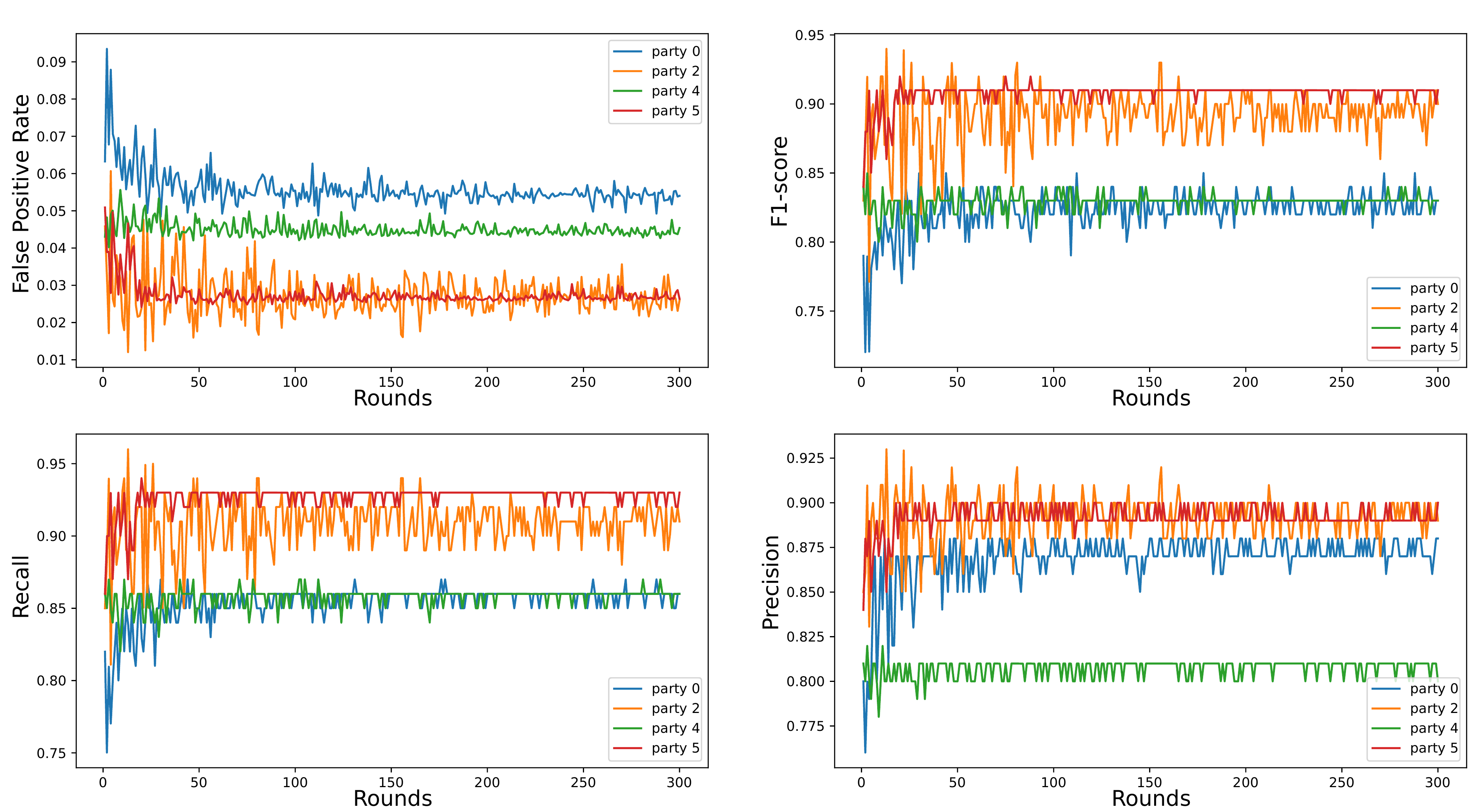}
    \caption{Mixed scenario's precision, recall, F-1 score and FPR with Fed+}
    \label{fig:Metrics_fp_mix}
\end{figure*}

Figure \ref{fig:Metrics_fp_mix} shows the evolution for the other metrics when using Fed+ with a similar trend as for the accuracy. Parties 2 and 4 have the best results for each metric: FPR=0.28, F1-score=0.91, recall=0.925 and precision=0.9. Parties 0 and 5 have similar results, except that party 0's precision is similar to parties 2 and 4. It should be noted that these results are similar to the balanced scenario. As in the previous case, it means that accuracy results are consistent with the values obtained for the other metrics.
 
Based on the obtained results, this scenario represents a trade-off between the previous two scenarios obtaining similar results to the balanced setting, where samples are shared among the different parties. Furthermore, previous results demonstrate the need for considering additional aggregation functions (beyond FedAvg) in order to deal with scenarios characterized by non-iid and skewed data among the parties that are common in real-world scenarios. 


\subsection{Comparison between basic, balanced, mixed, and distributed scenarios}
\begin{figure}[!htb]
    \centering
    \includegraphics[width=0.95\textwidth]{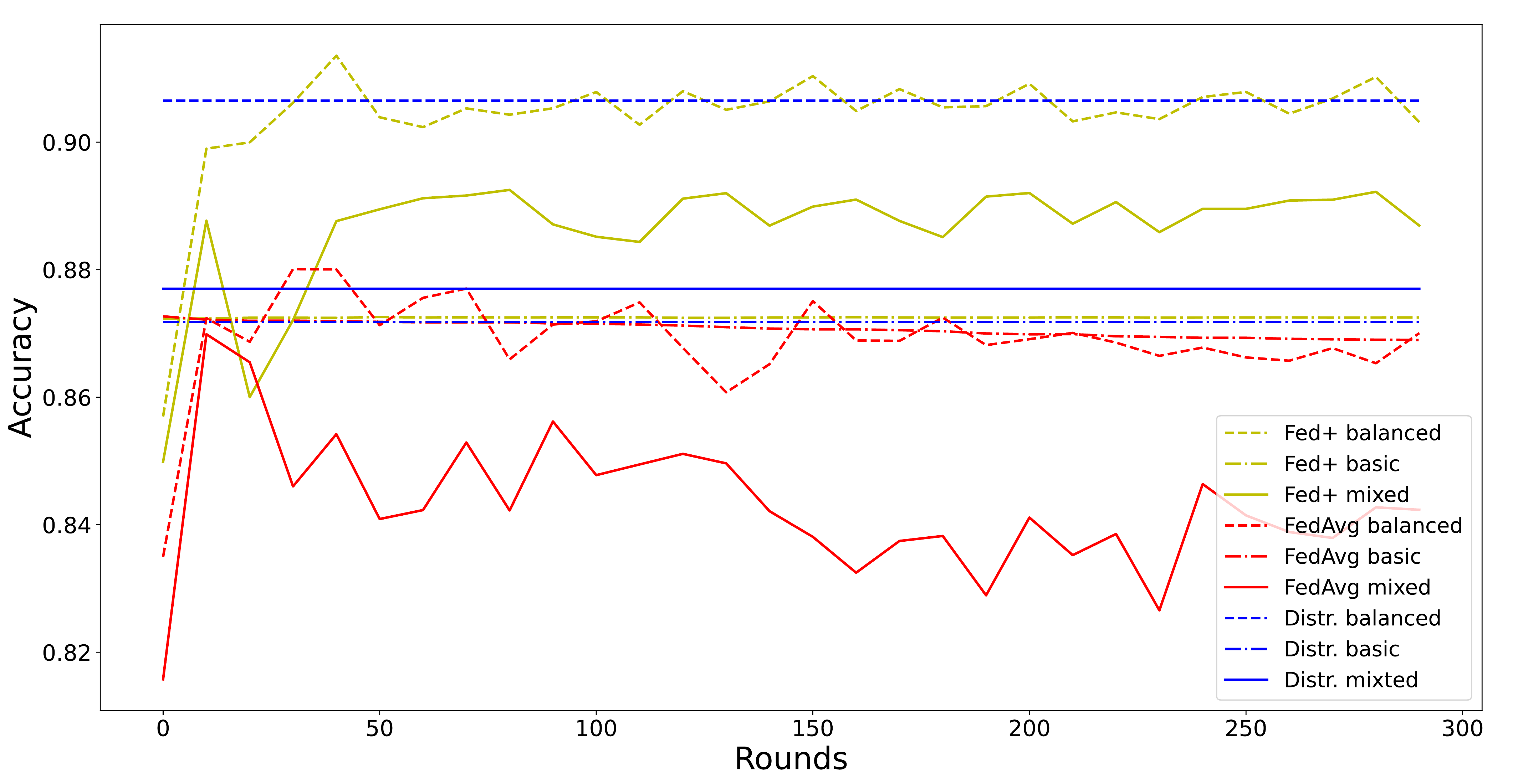}
    \caption{Comparison of average accuracy between basic, balanced, mixed and distributed scenarios.}
    \label{fig:todos_alg}
\end{figure}

After analyzing the different evaluation metrics, Figure \ref{fig:todos_alg} shows a comparison of the average accuracy of the parties for each federated scenario and a distributed setting, considering FedAvg and Fed+. According to these results, Fed+ provides higher accuracy for all the federated scenarios being considered. This demonstrates that it is able to handle better scenarios where parties do not have balanced datasets.   


For the basic scenario, graphs are similar for FedAvg and Fed+. However, it should be noted that, in the case of Fed+, accuracy remains constant about 0.8725, which is close to the 0.8718 of the distributed method, whereas it drops slowly from 0.8725 when using FedAvg. In the balanced scenario, the initial accuracy starts at 0.8569 and rapidly grows to 0.9039 (where it remains stable throughout the rounds) for Fed+. When FedAvg is used, accuracy grows from 0.8349, until 0.88 after 50 rounds, but it gradually drops to 0.87. Compared with the distributed setting, Fed+ has a similar accuracy to the 0.9065 of such scenario, since all parties have the same amount of data and number of classes. However, FedAvg does not reach the accuracy of the distributed case. 
The main reason is that, while parties' datasets are balanced among each other, each local dataset is unbalanced in relation to the number of samples for each class.

In the case of the mixed scenario, accuracy (when Fed+ is used) goes from 0.8498 until 0.8876 after 50 rounds, and it remains stable until it finishes with 0.8869. Indeed, after about 40 rounds, Fed+ overtakes the accuracy for the distributed case (0.877). However, the behavior of FedAvg is worse than the distributed case. In particular, accuracy goes from 0.8157 to 0.8698 after 10 rounds, but then, it decreases slowly until 0.8423. Therefore, in this scenario, Fed+ clearly improves the behavior of FedAvg. 

Based on the previous evaluation, Fed+ provides better results than FedAvg, which could introduce convergence issues in certain situations. Indeed, Fed+ provides better results for the mixed scenario compared to the results of the  balanced setting when FedAvg is used. Based on the results for the different scenarios, it should be noted that the impact of different data distributions is more clear in the case of Fed+ and the distributed setting, where the best results are obtained for the balanced scenarios, while the basic scenario provides the lowest value for accuracy. However, in the case of FedAvg the basic and balanced scenarios provide similar accuracy results, while the mixed scenario presents lower accuracy values. In any case, as already mentioned, the use of Fed+ has a clear impact in the results obtained for the different scenarios by improving the evaluation metrics' values when FedAvg is employed.

\section{Challenges and research directions}\label{sec:challenges}
Based on the evaluation results provided in the previous section and the analysis of the literature on the use of FL \cite{kairouz2019advances,mothukuri2020survey}, below we describe some of the main challenges and future research directions to be considered for the development of FL-enabled IDS in the scope of IoT scenarios. As described by \cite{kairouz2019advances}, it should be noted that many of the challenges associated with the use of FL in such context will require multidisciplinary approaches, including the application of privacy techniques, cryptography, distributed optimization, or information theory.



    


\subsection{Deploying FL on IoT devices}\label{sec:challenges_deploying}
While our work focuses on the impact of different data distributions on FL by using a simulated testbed, a significant set of challenges is derived from the deployment of a FL framework on real IoT devices. Indeed, as described by \cite{SotAResConsFL}, the computational requirements of well-known ML approaches might not be satisfied by constrained IoT devices in terms of memory, computing power and energy consumption. This aspect can be aggravated in the case of applying DL techniques, which require in general more computing resources than ML. To address such limitations, a current trend is the use of intermediate nodes at the network edge, so that the end devices send their data to these nodes acting as FL clients \cite{lim2020federated, ye2020edgefed}. For example, \cite{hei2020trusted} use intermediate entities (called RSPs) in charge of performing the local training in an FL setting. A similar approach is also proposed by \cite{AdapFLResCons}, which used an edge computing architecture to determine the aggregation frequency of the global model. However, it should be noted that sharing network traffic with these intermediate nodes to identity potential attacks can still pose privacy concerns. Other approaches consist of the reduction of the data that needs to be sent by segmenting and representing it \cite{gonzalez2018beats}, as well as by exploring feature selection \cite{mafarja2020augmented, gonzalez2019methodology}. Therefore, more efforts are needed to analyze the practical limitations of FL approaches in IoT scenarios, as well as the security and privacy implications derived from the use of edge computing architectures. In this context, a potential research direction is associated with the application of TinyML frameworks (e.g., TensorFlow Lite \cite{warden2019tinyml}) in FL scenarios, as recently described by \cite{mathur2021device}.

\subsection{Limitations of existing IDS-IoT datasets for FL}
As described in Section \ref{sec:related}, some of the existing FL-enabled IDS proposals for IoT are based on general network datasets, which do not consider IoT technologies and devices. Even though some datasets have recently been proposed for IoT scenarios, as described by \cite{rey2021federated}, some of them cannot be applied in an FL environment, since they do not provide data associated with different IP addresses or devices, in particular the IP destinations that can be identified as the parties of the FL environment in IDS. Furthermore, as described in Section \ref{sec:methodology}, most IDS datasets for IoT present a significant imbalance between benign and attack traffic, as well as a limited set of attacks being considered. Moreover, we note that ToN\_IoT is the only dataset that considers possible security threats related to telemetry data and sensor readings, unlike other datasets only dealing with network attacks. However, as described by \cite{zarpelao2017survey}, the development of IDS datasets for IoT still needs to consider a broader scope of IoT technologies (including well-known protocols like CoAP \cite{shelby2014constrained}), as well as additional aspects (e.g., energy consumption) that can serve to identify potential attacks. Therefore, more effort is needed in the development of IDS datasets for IoT considering its divisibility to be deployed in a FL setting.

\subsection{Aggregator as bottleneck}
Even though FL is based on a collaborative training approach, the coordinator entity may become a bottleneck from a performance and privacy perspective, as well as a single point of failure. To address such issue, a current trend is the application of blockchain technology \cite{zheng2018blockchain}, which represents a distributed and immutable ledger shared by several nodes. The use of blockchain can increase the level of trust in an FL environment, where the centralized coordinator is replaced by a set of nodes with distributed functionality, which is carried out through smart contracts. Indeed, blockchain has been proposed in recent works to make model updates accountable and avoid potentially malicious updates \cite{zhao2020privacy}. In the context of an IDS approach for IoT, \cite{hei2020trusted} uses intermediate nodes acting as blockchain clients to store the model parameters updated by the end devices to avoid potential manipulation. Despite these efforts, we note that most of current approaches do not provide comprehensive evaluations considering training frequency and scenarios with a large number of devices, which may be required for IDS approaches. Furthermore, as described by \cite{kairouz2019advances}, the use of permissionless blockchains (e.g., Ethereum \cite{wood2014ethereum}) can raise privacy concerns, which must be addressed by proper encryption or differential privacy techniques, as described in Section \ref{sec:challenges_privacy}.

\subsection{Communication requirements}
The need for a significant communication bandwidth to exchange global model updates represents a well-known issue associated with the use of FL \cite{SotAResConsFL}. This problem can be exacerbated in IoT scenarios where end devices acting as FL clients need to communicate their model updates through constrained networks and devices, which can degrade the network or IoT performance \cite{hu2020differentially}. In general, there are two main factors that impose strong communication requirements between FL clients and coordinator. The first aspect is related to the amount of data associated with the gradient exchange \cite{hamer2020fedboost}, which is required between clients and the coordinator for the learning process. This is generally addressed by gradient compression techniques, such as quantization and sparsification, as described by \cite{liu2020communication}. The second aspect is related to the number of training rounds required to converge the model that can vary depending on the scenario, dataset, data distribution, or the ML algorithm being considered. For example, based on our evaluation results, the different metrics remain stable after 50 rounds in the balanced and mixed scenarios, although this may be different with other evaluation conditions. While a common trend to reduce the training rounds is to perform several local training iterations before updating the global model \cite{guha2019one}, the execution of such local training iterations may have a significant impact on FL clients, specially in case of resource-constrained devices (see Section \ref{sec:challenges_deploying}). 

\subsection{Client selection}
As described in Section \ref{sec:background}, in each training round, the coordinator can select a subset of devices to participate as FL clients in the training process. For this purpose, different aspects such as device status, battery level, computing/communication capacity, or ML technique's accuracy could be considered \cite{abdulrahman2020fedmccs, mohammed2020budgeted}. Indeed, the client selection process can have an impact on the obtained accuracy and, therefore, on the detection of potential security attacks in the scope of an IDS approach. In our case, according to the results described in Section \ref{sec:evaluation}, we found that even a static client selection process can help to obtaining a better performance of the ML algorithm. However, more sophisticated client selection strategies must consider the dynamic aspects of an IoT environment in each training round. For example, some devices may not be available in a certain round due to mobility issues or loss of connectivity \cite{lim2020federated}. Furthermore, due to devices heterogeneity, while some of them could perform the local training in a few milliseconds, other devices could require a longer period to update the model (e.g., due to resource constraints), which could slow down the overall federated training \cite{nishio2019client}. In the context of IDS, this could lead to a longer delay in detecting a certain attack, which could have severe consequences on the overall cybersecurity of the network. An additional aspect is related to the need to provide incentives to devices, in order to foster their participation in the training process \cite{rahman2020survey}. Otherwise, some devices may not want to use their limited resources for this purpose. While some recent works address this issue in IoT scenarios \cite{zhan2020learning}, more efforts are required in real IoT environments to evaluate its impact on the learning process.

\subsection{Dynamic IoT devices' behavior throughout their lifecycle}
Another aspect is the need to consider the changing behavior of IoT devices throughout their lifecycle. For example, a software update process for a certain device can change its behavior \cite{hernandez2020updating}, so that a new learning process is required in order to reflect the new behavior as \textit{benign traffic} in the context of an IDS. However, such change could be also related to a potential attach affecting this device. Therefore, there is a need to integrate network management approaches to detect if behavioral changes in a certain device are produced intentionally, or they are due to a malicious action. Furthermore, the behavioral changes of a single device could affect to the behavior of other interacting devices. In the case of a FL scenario, it could require new training rounds that might have a significant impact specially in settings with constrained devices and networks. However, this aspect is not addressed  existing FL-enabled IDS approaches, which are based on existing datasets that do not reflect potential behavioral changes on IoT devices throughout their lifecycle. 

\subsection{Security attacks}
Like in the case of centralized approaches, FL is also susceptible to several attacks that can affect the learning process. Indeed, as described in recent works \cite{rahman2020survey}, some of the major security threats in FL are represented by \textit{data poisoning} and \textit{model update poisoning} attacks. The former is related to the attacker ability to add false training data or modify the existing dataset of a certain client, for example, by modifying the labels (label-flipping). The latter focuses on changing the global model instead of the local training dataset. The realization of such attacks could cause false alarms in an IDS approach due to misclassification of benign/malicious traffic \cite{nguyen2020poisoning}. To address such concerns, a recent work evaluates the behavior of different aggregation functions against several security attacks in an FL-enabled IDS approach \cite{rey2021federated}. Indeed, the application of certain aggregation approaches could help to make an FL setting more robust against potential attacks. In this direction, as part of our future work, we will evaluate how Fed+ behaves in the context of different data poisoning and model update poisoning attacks with different data distributions. Other complementary approaches to be considered are based on network management approaches to ensure that only devices behaving as intended can participate in the training process \cite{feraudo2020colearn}. These proposals still require lightweight cryptographic mechanisms to be considered in real IoT environments. Additionally, trust and reputation mechanisms can also be used in order to prevent malicious nodes from injecting false data into the training phase, even when using suitable cryptographic approaches \cite{kang2020reliable}.

\subsection{Privacy concerns} \label{sec:challenges_privacy}
While FL was mainly proposed to mitigate the privacy concerns associated with centralized learning approaches, it can still leak information from clients' training data. Indeed, as described by \cite{rahman2020survey}, a malicious server could infer information from model updates, as well as alter them in order to fool the global model. This can be exacerbated in the context of IDS approaches to IoT, where device network traffic data can reveal everyday user habits. Therefore, the application of privacy-preserving techniques for FL has attracted a significant interest recently \cite{mothukuri2020survey}, including the use of differential privacy (DP) approaches \cite{wei2020federated}, secure-multiparty computation (SMC) \cite{zhao2019secure} and homomorphic encryption \cite{zhang2020batchcrypt}. However, these techniques often come at a cost in terms of accuracy and efficiency \cite{FLChalMethFutDir, mothukuri2020survey}, which can negatively affect the attack detection capabilities of IDS approaches. Indeed, a recent work evaluates the application of DP for an FL-enabled IDS considering non-iid data \cite{FLHeteCohort}. Although other recent efforts have been proposed for IoT scenarios \cite{hu2020differentially}, more studies are required to come up with a tradeoff between privacy requirements, as well as performance and accuracy requirements for effective IDS approaches.



\section{Conclusions}\label{sec:conclusions}
The application of FL techniques has attracted a significant interest in recent years due to their advantages over traditional centralized learning approaches. In this work, we provided an overview about the current research efforts for the application of FL toward the development of IDS approaches for IoT scenarios. Unlike previous works, we considered several settings with different data distributions. Our evaluation demonstrates the impact of non-iid and highly skewed data distributions on the FL performance, which directly affects the effectiveness of the security attack detection. We demonstrate that an instance selection process based on the Shannon entropy of each local dataset can improve the overall accuracy obtaining similar results compared with a scenario where the dataset is balanced among the parties. Toward this end, we evaluated the use of the FedAvg and Fed+ aggregation functions using the recently proposed ToN\_IoT dataset. Furthermore, based on our evaluation and the analysis of existing literature, we described the main challenges to be considered in the coming years for the deployment of FL-enabled IDS in IoT. As future work, we will address some of such challenges by deploying a FL-enabled IDS approach in real IoT scenarios to assess its feasibility in environments with constrained devices and networks. Furthermore, we will analyze the potential application of personalized FL, where each node uses the most appropriate learning model, in order to improve the overall accuracy for attack detection in IoT scenarios.

\section*{Acknowledgment}
This work has been sponsored by UMU-CAMPUS LIVING LAB EQC2019-006176-P funded by ERDF funds, by the European Commission through the PHOENIX (grant agreement  893079) CyberSec4Europe (g.a. 830929) and DEMETER (g.a. 857202) EU Projects. It was also co-financed by the European Social Fund (ESF) and the Youth European Initiative (YEI) under the Spanish Seneca Foundation (CARM).

\bibliographystyle{elsarticle-num}
\bibliography{biblio}

\begin{thebibliography}{10}
\expandafter\ifx\csname url\endcsname\relax
  \def\url#1{\texttt{#1}}\fi
\expandafter\ifx\csname urlprefix\endcsname\relax\def\urlprefix{URL }\fi
\expandafter\ifx\csname href\endcsname\relax
  \def\href#1#2{#2} \def\path#1{#1}\fi

\bibitem{neshenko2019demystifying}
N.~Neshenko, E.~Bou-Harb, J.~Crichigno, G.~Kaddoum, N.~Ghani, Demystifying
  {IoT} security: an exhaustive survey on {IoT} vulnerabilities and a first
  empirical look on internet-scale {IoT} exploitations, IEEE Communications
  Surveys \& Tutorials 21~(3) (2019) 2702--2733.

\bibitem{pour2020data}
M.~S. Pour, A.~Mangino, K.~Friday, M.~Rathbun, E.~Bou-Harb, F.~Iqbal,
  S.~Samtani, J.~Crichigno, N.~Ghani, On data-driven curation, learning, and
  analysis for inferring evolving internet-of-things (iot) botnets in the wild,
  Computers \& Security 91 (2020) 101707.

\bibitem{da2019internet}
K.~A. da~Costa, J.~P. Papa, C.~O. Lisboa, R.~Munoz, V.~H.~C. de~Albuquerque,
  Internet of things: A survey on machine learning-based intrusion detection
  approaches, Computer Networks 151 (2019) 147--157.

\bibitem{ding2019survey}
W.~Ding, X.~Jing, Z.~Yan, L.~T. Yang, A survey on data fusion in internet of
  things: Towards secure and privacy-preserving fusion, Information Fusion 51
  (2019) 129--144.

\bibitem{iggena2021iotcrawler}
T.~Iggena, E.~Bin~Ilyas, M.~Fischer, R.~T{\"o}njes, T.~Elsaleh, R.~Rezvani,
  N.~Pourshahrokhi, S.~Bischof, A.~Fernbach, J.~Xavier~Parreira, et~al.,
  Iotcrawler: Challenges and solutions for searching the internet of things,
  Sensors 21~(5) (2021) 1559.

\bibitem{mcmahan2017communication}
B.~McMahan, E.~Moore, D.~Ramage, S.~Hampson, B.~A. y~Arcas,
  Communication-efficient learning of deep networks from decentralized data,
  in: Artificial Intelligence and Statistics, PMLR, 2017, pp. 1273--1282.

\bibitem{zhao2018federated}
Y.~Zhao, M.~Li, L.~Lai, N.~Suda, D.~Civin, V.~Chandra, Federated learning with
  non-iid data, arXiv preprint arXiv:1806.00582 (2018).

\bibitem{DIOT}
T.~D. Nguyen, S.~Marchal, M.~Miettinen, H.~Fereidooni, N.~Asokan, A.-R.
  Sadeghi, D{\"i}ot: A federated self-learning anomaly detection system for
  iot, in: 2019 IEEE 39th International Conference on Distributed Computing
  Systems (ICDCS), IEEE, 2019, pp. 756--767.

\bibitem{FLMimicIDS}
N.~A. A.-A. Al-Marri, B.~S. Ciftler, M.~M. Abdallah, Federated mimic learning
  for privacy preserving intrusion detection, arXiv:2012.06974v1 (2020).

\bibitem{LocKedge}
T.~Huong, P.~B. Ta, D.~Long, B.~Thang, N.~Binh, T.~Luong, K.~P. TRAN, Lockedge:
  Low-complexity cyberattack detection in iot edge computing, IEEE Access PP
  (2021).

\bibitem{BFL-CIDS}
X.~Hei, X.~Yin, Y.~Wang, J.~Ren, L.~Zhu, A trusted feature aggregator federated
  learning for distributed malicious attack detection, Computers \& Security 99
  (2020) 102033.

\bibitem{rey2021federated}
V.~Rey, P.~M.~S. S{\'a}nchez, A.~H. Celdr{\'a}n, G.~Bovet, M.~Jaggi, Federated
  learning for malware detection in iot devices, arXiv preprint
  arXiv:2104.09994 (2021).

\bibitem{booij2021ton_iot}
T.~M. Booij, I.~Chiscop, E.~Meeuwissen, N.~Moustafa, F.~T. den Hartog,
  Ton\_iot: The role of heterogeneity and the need for standardization of
  features and attack types in iot network intrusion datasets, IEEE Internet of
  Things Journal (2021).

\bibitem{alsaedi2020ton_iot}
A.~Alsaedi, N.~Moustafa, Z.~Tari, A.~Mahmood, A.~Anwar, Ton\_iot telemetry
  dataset: a new generation dataset of iot and iiot for data-driven intrusion
  detection systems, IEEE Access 8 (2020) 165130--165150.

\bibitem{bonachela2008entropy}
J.~A. Bonachela, H.~Hinrichsen, M.~A. Munoz, Entropy estimates of small data
  sets, Journal of Physics A: Mathematical and Theoretical 41~(20) (2008)
  202001.

\bibitem{datasets}
\href{https://github.com/Enrique-Marmol/Evaluating-FL-for-Intrusion-Detection-in-IoT-review-and-challenges}{Evaluating-fl-for-intrusion-detection-in-iot-review-and-challenges
  datasets} (2021).
\newline\urlprefix\url{https://github.com/Enrique-Marmol/Evaluating-FL-for-Intrusion-Detection-in-IoT-review-and-challenges}

\bibitem{li2019convergence}
X.~Li, K.~Huang, W.~Yang, S.~Wang, Z.~Zhang, On the convergence of fedavg on
  non-iid data, arXiv preprint arXiv:1907.02189 (2019).

\bibitem{fed+paper}
P.~Yu, L.~Wynter, S.~H. Lim, Fed+: A family of fusion algorithms for federated
  learning (2020).
\newblock \href {http://arxiv.org/abs/2009.06303} {\path{arXiv:2009.06303}}.

\bibitem{ludwig2020ibm}
H.~Ludwig, N.~Baracaldo, G.~Thomas, Y.~Zhou, A.~Anwar, S.~Rajamoni, Y.~Ong,
  J.~Radhakrishnan, A.~Verma, M.~Sinn, et~al., Ibm federated learning: an
  enterprise framework white paper v0. 1, arXiv preprint arXiv:2007.10987
  (2020).

\bibitem{khraisat2019survey}
A.~Khraisat, I.~Gondal, P.~Vamplew, J.~Kamruzzaman, Survey of intrusion
  detection systems: techniques, datasets and challenges, Cybersecurity 2~(1)
  (2019) 1--22.

\bibitem{chapaneri2019comprehensive}
R.~Chapaneri, S.~Shah, A comprehensive survey of machine learning-based network
  intrusion detection, Smart Intelligent Computing and Applications (2019)
  345--356.

\bibitem{liu2019machine}
H.~Liu, B.~Lang, Machine learning and deep learning methods for intrusion
  detection systems: A survey, applied sciences 9~(20) (2019) 4396.

\bibitem{yin2017deep}
C.~Yin, Y.~Zhu, J.~Fei, X.~He, A deep learning approach for intrusion detection
  using recurrent neural networks, Ieee Access 5 (2017) 21954--21961.

\bibitem{drewek2021survey}
A.~Drewek-Ossowicka, M.~Pietro{\l}aj, J.~Rumi{\'n}ski, A survey of neural
  networks usage for intrusion detection systems, Journal of Ambient
  Intelligence and Humanized Computing 12 (2021) 497--514.

\bibitem{liang2019industrial}
W.~Liang, K.-C. Li, J.~Long, X.~Kui, A.~Y. Zomaya, An industrial network
  intrusion detection algorithm based on multifeature data clustering
  optimization model, IEEE Transactions on Industrial Informatics 16~(3) (2019)
  2063--2071.

\bibitem{ferrag2021deep}
M.~A. Ferrag, L.~Shu, H.~Djallel, K.-K.~R. Choo, Deep learning-based intrusion
  detection for distributed denial of service attack in agriculture 4.0,
  Electronics 10~(11) (2021) 1257.

\bibitem{ge2021towards}
M.~Ge, N.~F. Syed, X.~Fu, Z.~Baig, A.~Robles-Kelly, Towards a deep
  learning-driven intrusion detection approach for internet of things, Computer
  Networks 186 (2021) 107784.

\bibitem{garcia2021distributed}
N.~Garcia, T.~Alcaniz, A.~Gonz{\'a}lez-Vidal, J.~B. Bernabe, D.~Rivera,
  A.~Skarmeta, Distributed real-time slowdos attacks detection over encrypted
  traffic using artificial intelligence, Journal of Network and Computer
  Applications 173 (2021) 102871.

\bibitem{rahman2020internet}
S.~A. Rahman, H.~Tout, C.~Talhi, A.~Mourad, Internet of things intrusion
  detection: Centralized, on-device, or federated learning?, IEEE Network
  34~(6) (2020) 310--317.

\bibitem{eskandari2020passban}
M.~Eskandari, Z.~H. Janjua, M.~Vecchio, F.~Antonelli, Passban ids: an
  intelligent anomaly-based intrusion detection system for iot edge devices,
  IEEE Internet of Things Journal 7~(8) (2020) 6882--6897.

\bibitem{rahman2020survey}
S.~A. Rahman, H.~Tout, H.~Ould-Slimane, A.~Mourad, C.~Talhi, M.~Guizani, A
  survey on federated learning: The journey from centralized to distributed
  on-site learning and beyond, IEEE Internet of Things Journal (2020).

\bibitem{nishio2019client}
T.~Nishio, R.~Yonetani, Client selection for federated learning with
  heterogeneous resources in mobile edge, in: ICC 2019-2019 IEEE International
  Conference on Communications (ICC), IEEE, 2019, pp. 1--7.

\bibitem{li2018federated}
T.~Li, A.~K. Sahu, M.~Zaheer, M.~Sanjabi, A.~Talwalkar, V.~Smith, Federated
  optimization in heterogeneous networks, arXiv preprint arXiv:1812.06127
  (2018).

\bibitem{mothukuri2020survey}
V.~Mothukuri, R.~M. Parizi, S.~Pouriyeh, Y.~Huang, A.~Dehghantanha,
  G.~Srivastava, A survey on security and privacy of federated learning, Future
  Generation Computer Systems 115 (2020) 619--640.

\bibitem{feraudo2020colearn}
A.~Feraudo, P.~Yadav, V.~Safronov, D.~A. Popescu, R.~Mortier, S.~Wang,
  P.~Bellavista, J.~Crowcroft, Colearn: Enabling federated learning in
  mud-compliant iot edge networks, in: Proceedings of the Third ACM
  International Workshop on Edge Systems, Analytics and Networking, 2020, pp.
  25--30.

\bibitem{nguyen2021federated}
D.~C. Nguyen, M.~Ding, P.~N. Pathirana, A.~Seneviratne, J.~Li, H.~V. Poor,
  Federated learning for internet of things: A comprehensive survey, arXiv
  preprint arXiv:2104.07914 (2021).

\bibitem{FLEAM}
J.~Li, L.~Lyu, X.~Liu, X.~Zhang, X.~Lyu, Fleam: A federated learning empowered
  architecture to mitigate ddos in industrial iot, arXiv:2012.06150 (2020).

\bibitem{FLHeteCohort}
A.~K. Chathoth, A.~Jagannatha, S.~Lee, Federated intrusion detection for iot
  with heterogeneous cohort privacy, arXiv:2101.09878v1 (2021).

\bibitem{BNNSIDFL}
Q.~{Qin}, K.~{Poularakis}, K.~K. {Leung}, L.~{Tassiulas}, Line-speed and
  scalable intrusion detection at the network edge via federated learning
  (2020) 352--360.

\bibitem{CLCAIoT40}
T.~V. {Khoa}, Y.~M. {Saputra}, D.~T. {Hoang}, N.~L. {Trung}, D.~{Nguyen}, N.~V.
  {Ha}, E.~{Dutkiewicz}, Collaborative learning model for cyberattack detection
  systems in iot industry 4.0 (2020) 1--6\href
  {https://doi.org/10.1109/WCNC45663.2020.9120761}
  {\path{doi:10.1109/WCNC45663.2020.9120761}}.

\bibitem{fediotguard}
V.~Rey, \href{https://github.com/ValerianRey/fed\_iot\_guard}{fed\_iot\_guard}.
\newline\urlprefix\url{https://github.com/ValerianRey/fed\_iot\_guard}

\bibitem{DEEPFED}
B.~Li, Y.~Wu, J.~Song, R.~Lu, T.~Li, L.~Zhao, Deepfed: Federated deep learning
  for intrusiondetection in industrial cyber-physical systems, IEEE (2020).

\bibitem{morris2014industrial}
T.~Morris, W.~Gao, Industrial control system traffic data sets for intrusion
  detection research, in: International Conference on Critical Infrastructure
  Protection, Springer, 2014, pp. 65--78.

\bibitem{grinberg2021flask}
M.~Grinberg, Flask-socketio documentation (2021).

\bibitem{ketkar2017introduction}
N.~Ketkar, Introduction to keras, in: Deep learning with Python, Springer,
  2017, pp. 97--111.

\bibitem{mothukuri2021federated}
V.~Mothukuri, P.~Khare, R.~M. Parizi, S.~Pouriyeh, A.~Dehghantanha,
  G.~Srivastava, Federated learning-based anomaly detection for iot security
  attacks, IEEE Internet of Things Journal (2021).

\bibitem{7348942}
N.~{Moustafa}, J.~{Slay}, Unsw-nb15: a comprehensive data set for network
  intrusion detection systems (unsw-nb15 network data set), in: 2015 Military
  Communications and Information Systems Conference (MilCIS), 2015, pp. 1--6.
\newblock \href {https://doi.org/10.1109/MilCIS.2015.7348942}
  {\path{doi:10.1109/MilCIS.2015.7348942}}.

\bibitem{dey2017gate}
R.~Dey, F.~M. Salem, Gate-variants of gated recurrent unit (gru) neural
  networks, in: 2017 IEEE 60th international midwest symposium on circuits and
  systems (MWSCAS), IEEE, 2017, pp. 1597--1600.

\bibitem{antonakakis2017understanding}
M.~Antonakakis, T.~April, M.~Bailey, M.~Bernhard, E.~Bursztein, J.~Cochran,
  Z.~Durumeric, J.~A. Halderman, L.~Invernizzi, M.~Kallitsis, et~al.,
  Understanding the mirai botnet, in: 26th $\{$USENIX$\}$ security symposium
  ($\{$USENIX$\}$ Security 17), 2017, pp. 1093--1110.

\bibitem{stolfo1999kdd}
S.~Stolfo, S.~Stolfo, Kdd cup 1999 dataset, UCI KDD repository. http://kdd.
  ics. uci. edu (1999).

\bibitem{tavallaee2009detailed}
M.~Tavallaee, E.~Bagheri, W.~Lu, A.~A. Ghorbani, A detailed analysis of the kdd
  cup 99 data set, in: 2009 IEEE symposium on computational intelligence for
  security and defense applications, IEEE, 2009, pp. 1--6.

\bibitem{wei2020federated}
K.~Wei, J.~Li, M.~Ding, C.~Ma, H.~H. Yang, F.~Farokhi, S.~Jin, T.~Q. Quek,
  H.~V. Poor, Federated learning with differential privacy: Algorithms and
  performance analysis, IEEE Transactions on Information Forensics and Security
  15 (2020) 3454--3469.

\bibitem{sharafaldin2018toward}
I.~Sharafaldin, A.~H. Lashkari, A.~A. Ghorbani, Toward generating a new
  intrusion detection dataset and intrusion traffic characterization., in:
  ICISSp, 2018, pp. 108--116.

\bibitem{hubara2016binarized}
I.~Hubara, M.~Courbariaux, D.~Soudry, R.~El-Yaniv, Y.~Bengio, Binarized neural
  networks, in: Proceedings of the 30th International Conference on Neural
  Information Processing Systems, 2016, pp. 4114--4122.

\bibitem{beigi2014towards}
E.~B. Beigi, H.~H. Jazi, N.~Stakhanova, A.~A. Ghorbani, Towards effective
  feature selection in machine learning-based botnet detection approaches, in:
  2014 IEEE Conference on Communications and Network Security, IEEE, 2014, pp.
  247--255.

\bibitem{bernstein2018signsgd}
J.~Bernstein, Y.-X. Wang, K.~Azizzadenesheli, A.~Anandkumar, signsgd:
  Compressed optimisation for non-convex problems, in: International Conference
  on Machine Learning, PMLR, 2018, pp. 560--569.

\bibitem{hinton2009deep}
G.~E. Hinton, Deep belief networks, Scholarpedia 4~(5) (2009) 5947.

\bibitem{meidan2018n}
Y.~Meidan, M.~Bohadana, Y.~Mathov, Y.~Mirsky, A.~Shabtai, D.~Breitenbacher,
  Y.~Elovici, N-baiot—network-based detection of iot botnet attacks using
  deep autoencoders, IEEE Pervasive Computing 17~(3) (2018) 12--22.

\bibitem{yin2018byzantine}
D.~Yin, Y.~Chen, R.~Kannan, P.~Bartlett, Byzantine-robust distributed learning:
  Towards optimal statistical rates, in: International Conference on Machine
  Learning, PMLR, 2018, pp. 5650--5659.

\bibitem{koroniotis2019towards}
N.~Koroniotis, N.~Moustafa, E.~Sitnikova, B.~Turnbull, Towards the development
  of realistic botnet dataset in the internet of things for network forensic
  analytics: Bot-iot dataset, Future Generation Computer Systems 100 (2019)
  779--796.

\bibitem{ToNIoTPaper}
A.~Alsaedi, N.~Moustafa, Z.~Tari, A.~Mahmood, A.~Anwar, Ton\_iot telemetry
  dataset: A new generation dataset of iot and iiot for data-driven intrusion
  detection systems, IEEE Access 8 (2020) 165130--165150.
\newblock \href {https://doi.org/10.1109/ACCESS.2020.3022862}
  {\path{doi:10.1109/ACCESS.2020.3022862}}.

\bibitem{guerra2020medbiot}
A.~Guerra-Manzanares, J.~Medina-Galindo, H.~Bahsi, S.~N{\~o}mm, Medbiot:
  Generation of an iot botnet dataset in a medium-sized iot network., in:
  ICISSP, 2020, pp. 207--218.

\bibitem{ullah2020scheme}
I.~Ullah, Q.~H. Mahmoud, A scheme for generating a dataset for anomalous
  activity detection in iot networks, in: Canadian Conference on Artificial
  Intelligence, Springer, 2020, pp. 508--520.

\bibitem{cictoniot}
\href{https://staff.itee.uq.edu.au/marius/NIDS\_datasets/}{Machine
  learning-based nids datasets}.
\newline\urlprefix\url{https://staff.itee.uq.edu.au/marius/NIDS\_datasets/}

\bibitem{lashkari2017characterization}
A.~H. Lashkari, G.~Draper-Gil, M.~S.~I. Mamun, A.~A. Ghorbani, Characterization
  of tor traffic using time based features., in: ICISSp, 2017, pp. 253--262.

\bibitem{bohning1992multinomial}
D.~B{\"o}hning, Multinomial logistic regression algorithm, Annals of the
  institute of Statistical Mathematics 44~(1) (1992) 197--200.

\bibitem{LogReg}
\href{https://towardsdatascience.com/logistic-regression-explained-9ee73cede081}{Logistic
  regression explained}.
\newline\urlprefix\url{https://towardsdatascience.com/logistic-regression-explained-9ee73cede081}

\bibitem{HeteInFedAvg}
J.~Pang, Y.~Huang, Z.~Xie, Q.~Han, Z.~Cai, Realizing the heterogeneity: A
  self-organized federated learning framework for iot, IEEE Internet of Things
  Journal 8~(5) (2021) 3088--3098.
\newblock \href {https://doi.org/10.1109/JIOT.2020.3007662}
  {\path{doi:10.1109/JIOT.2020.3007662}}.

\bibitem{kairouz2019advances}
P.~Kairouz, H.~B. McMahan, B.~Avent, A.~Bellet, M.~Bennis, A.~N. Bhagoji,
  K.~Bonawitz, Z.~Charles, G.~Cormode, R.~Cummings, et~al., Advances and open
  problems in federated learning, arXiv preprint arXiv:1912.04977 (2019).

\bibitem{SotAResConsFL}
A.~Imteaj, U.~Thakker, S.~Wang, J.~Li, M.~H. Amini, Federated learning for
  resource-constrained iot devices:panoramas and state-of-the-art,
  arXiv:2002.10610v1 (2020).

\bibitem{lim2020federated}
W.~Y.~B. Lim, N.~C. Luong, D.~T. Hoang, Y.~Jiao, Y.-C. Liang, Q.~Yang,
  D.~Niyato, C.~Miao, Federated learning in mobile edge networks: A
  comprehensive survey, IEEE Communications Surveys \& Tutorials 22~(3) (2020)
  2031--2063.

\bibitem{ye2020edgefed}
Y.~Ye, S.~Li, F.~Liu, Y.~Tang, W.~Hu, Edgefed: optimized federated learning
  based on edge computing, IEEE Access 8 (2020) 209191--209198.

\bibitem{hei2020trusted}
X.~Hei, X.~Yin, Y.~Wang, J.~Ren, L.~Zhu, A trusted feature aggregator federated
  learning for distributed malicious attack detection, Computers \& Security 99
  (2020) 102033.

\bibitem{AdapFLResCons}
S.~Wang, T.~Tuor, T.~Salonidis, K.~K. Leung, C.~Makaya, T.~He, K.~Chan,
  Adaptive federated learning in resource constrained edge computing systems,
  arXiv:1804.05271v3 (2019).

\bibitem{gonzalez2018beats}
A.~Gonzalez-Vidal, P.~Barnaghi, A.~F. Skarmeta, Beats: Blocks of eigenvalues
  algorithm for time series segmentation, IEEE Transactions on Knowledge and
  Data Engineering 30~(11) (2018) 2051--2064.

\bibitem{mafarja2020augmented}
M.~Mafarja, A.~A. Heidari, M.~Habib, H.~Faris, T.~Thaher, I.~Aljarah, Augmented
  whale feature selection for iot attacks: Structure, analysis and
  applications, Future Generation Computer Systems 112 (2020) 18--40.

\bibitem{gonzalez2019methodology}
A.~Gonzalez-Vidal, F.~Jimenez, A.~F. Gomez-Skarmeta, A methodology for energy
  multivariate time series forecasting in smart buildings based on feature
  selection, Energy and Buildings 196 (2019) 71--82.

\bibitem{warden2019tinyml}
P.~Warden, D.~Situnayake, Tinyml: Machine learning with tensorflow lite on
  arduino and ultra-low-power microcontrollers, " O'Reilly Media, Inc.", 2019.

\bibitem{mathur2021device}
A.~Mathur, D.~J. Beutel, P.~P.~B. de~Gusm{\~a}o, J.~Fernandez-Marques,
  T.~Topal, X.~Qiu, T.~Parcollet, Y.~Gao, N.~D. Lane, On-device federated
  learning with flower, arXiv preprint arXiv:2104.03042 (2021).

\bibitem{zarpelao2017survey}
B.~B. Zarpel{\~a}o, R.~S. Miani, C.~T. Kawakani, S.~C. de~Alvarenga, A survey
  of intrusion detection in internet of things, Journal of Network and Computer
  Applications 84 (2017) 25--37.

\bibitem{shelby2014constrained}
Z.~Shelby, K.~Hartke, C.~Bormann, The constrained application protocol (coap)
  (2014).

\bibitem{zheng2018blockchain}
Z.~Zheng, S.~Xie, H.-N. Dai, X.~Chen, H.~Wang, Blockchain challenges and
  opportunities: A survey, International Journal of Web and Grid Services
  14~(4) (2018) 352--375.

\bibitem{zhao2020privacy}
Y.~Zhao, J.~Zhao, L.~Jiang, R.~Tan, D.~Niyato, Z.~Li, L.~Lyu, Y.~Liu,
  Privacy-preserving blockchain-based federated learning for iot devices, IEEE
  Internet of Things Journal (2020).

\bibitem{wood2014ethereum}
G.~Wood, et~al., Ethereum: A secure decentralised generalised transaction
  ledger, Ethereum project yellow paper 151~(2014) (2014) 1--32.

\bibitem{hu2020differentially}
R.~Hu, Y.~Guo, E.~P. Ratazzi, Y.~Gong, Differentially private federated
  learning for resource-constrained internet of things, arXiv preprint
  arXiv:2003.12705 (2020).

\bibitem{hamer2020fedboost}
J.~Hamer, M.~Mohri, A.~T. Suresh, Fedboost: A communication-efficient algorithm
  for federated learning, in: International Conference on Machine Learning,
  PMLR, 2020, pp. 3973--3983.

\bibitem{liu2020communication}
Y.~Liu, N.~Kumar, Z.~Xiong, W.~Y.~B. Lim, J.~Kang, D.~Niyato,
  Communication-efficient federated learning for anomaly detection in
  industrial internet of things, in: GLOBECOM, Vol. 2020, 2020, pp. 1--6.

\bibitem{guha2019one}
N.~Guha, A.~Talwalkar, V.~Smith, One-shot federated learning, arXiv preprint
  arXiv:1902.11175 (2019).

\bibitem{abdulrahman2020fedmccs}
S.~AbdulRahman, H.~Tout, A.~Mourad, C.~Talhi, Fedmccs: multicriteria client
  selection model for optimal iot federated learning, IEEE Internet of Things
  Journal 8~(6) (2020) 4723--4735.

\bibitem{mohammed2020budgeted}
I.~Mohammed, S.~Tabatabai, A.~Al-Fuqaha, F.~El~Bouanani, J.~Qadir, B.~Qolomany,
  M.~Guizani, Budgeted online selection of candidate iot clients to participate
  in federated learning, IEEE Internet of Things Journal (2020).

\bibitem{zhan2020learning}
Y.~Zhan, P.~Li, Z.~Qu, D.~Zeng, S.~Guo, A learning-based incentive mechanism
  for federated learning, IEEE Internet of Things Journal 7~(7) (2020)
  6360--6368.

\bibitem{hernandez2020updating}
J.~L. Hern{\'a}ndez-Ramos, G.~Baldini, S.~N. Matheu, A.~Skarmeta, Updating iot
  devices: challenges and potential approaches, in: 2020 Global Internet of
  Things Summit (GIoTS), IEEE, 2020, pp. 1--5.

\bibitem{nguyen2020poisoning}
T.~D. Nguyen, P.~Rieger, M.~Miettinen, A.-R. Sadeghi, Poisoning attacks on
  federated learning-based iot intrusion detection system, in: NDSS Workshop on
  Decentralized IoT Systems and Security, 2020.

\bibitem{kang2020reliable}
J.~Kang, Z.~Xiong, D.~Niyato, Y.~Zou, Y.~Zhang, M.~Guizani, Reliable federated
  learning for mobile networks, IEEE Wireless Communications 27~(2) (2020)
  72--80.

\bibitem{zhao2019secure}
C.~Zhao, S.~Zhao, M.~Zhao, Z.~Chen, C.-Z. Gao, H.~Li, Y.-a. Tan, Secure
  multi-party computation: Theory, practice and applications, Information
  Sciences 476 (2019) 357--372.

\bibitem{zhang2020batchcrypt}
C.~Zhang, S.~Li, J.~Xia, W.~Wang, F.~Yan, Y.~Liu, Batchcrypt: Efficient
  homomorphic encryption for cross-silo federated learning, in: 2020 USENIX
  Annual Technical Conference, 2020, pp. 493--506.

\bibitem{FLChalMethFutDir}
T.~Li, A.~K. Sahu, A.~Talwalkar, V.~Smith, Federated learning:challenges,
  methods, and future directions, arXiv:1908.07873v1 (2019).

\end{thebibliography}

\end{document}